\documentclass[10pt,twocolumn,letterpaper]{article}

\usepackage[pagenumbers]{wacv} 

\usepackage{graphicx}
\usepackage{amsmath}
\usepackage{amssymb}
\usepackage{booktabs}

\usepackage{multirow}
\usepackage{color, colortbl}
\usepackage{amssymb}
\usepackage{pifont}
\newcommand{\xmark}{\ding{55}}%
\usepackage[dvipsnames]{xcolor}
\usepackage[export]{adjustbox}
\usepackage{subcaption}
\usepackage{arydshln}
\usepackage{lipsum}
\usepackage{enumitem}

\usepackage[pagebackref,breaklinks,colorlinks]{hyperref}

\usepackage[accsupp]{axessibility} 

\usepackage[capitalize]{cleveref}
\crefname{section}{Sec.}{Secs.}
\Crefname{section}{Section}{Sections}
\Crefname{table}{Table}{Tables}
\crefname{table}{Tab.}{Tabs.}

\def\wacvPaperID{2483} 
\def\confName{WACV}
\def\confYear{2025}

\begin{document}


\title{A Spatio-Temporal Representation Learning as an Alternative to Traditional Glosses in Sign Language Translation and Production}

\author{
Eui Jun Hwang\(^1\)\quad 
Sukmin Cho\(^1\)\quad 
Huije Lee\(^1\)\quad 
Youngwoo Yoon\(^2\)\quad 
Jong C. Park\(^1\)\thanks{Corresponding author.}\\
\(^1\)Korea Advanced Institute of Science and Technology,\\
\(^2\)Electronics and Telecommunications Research Institute\\
\(^1\){\tt\small \{ehwa20,nelllpic,angiquer,jongpark\}@kaist.ac.kr}\\
\(^2\){\tt\small youngwoo@etri.re.kr}
}
\maketitle

\begin{abstract}
This work addresses the challenges associated with the use of glosses in both Sign Language Translation (SLT) and Sign Language Production (SLP). While glosses have long been used as a bridge between sign language and spoken language, they come with two major limitations that impede the advancement of sign language systems. First, annotating the glosses is a labor-intensive and time-consuming process, which limits the scalability of datasets. Second, the glosses oversimplify sign language by stripping away its spatio-temporal dynamics, reducing complex signs to basic labels and missing the subtle movements essential for precise interpretation. To address these limitations, we introduce \textbf{Uni}versal \textbf{Glo}ss-level \textbf{R}epresentation (\textbf{UniGloR}), a framework designed to capture the spatio-temporal features inherent in sign language, providing a more dynamic and detailed alternative to the use of the glosses. The core idea of UniGloR is simple yet effective: We derive dense spatio-temporal representations from sign keypoint sequences using self-supervised learning and seamlessly integrate them into SLT and SLP tasks. Our experiments in a keypoint-based setting demonstrate that UniGloR either outperforms or matches the performance of previous SLT and SLP methods on two widely-used datasets: PHOENIX14T and How2Sign. 
Code is available at \url{https://github.com/eddie-euijun-hwang/UniGloR}.
\vspace{-1em}
\end{abstract}

\section{Introduction}

Sign language is a visual form of communication used by Deaf and Hard-of-Hearing communities. It utilizes a unique grammar that combines manual signals, such as hand movements and shapes, with non-manual signals, such as facial expressions, eye gaze, and body posture, to convey messages effectively. 
Moreover, there are over 300 distinct sign languages worldwide, used by more than 72 million Deaf or Hard-of-Hearing people\footnote{\scriptsize\url{https://education.nationalgeographic.org/resource/sign-language/}}. 
These factors highlight the importance of a bidirectional translation system that supports both \textit{sign-to-text} and \textit{text-to-sign} translations. Such a system is crucial for bridging the communication gap between hearing and Deaf communities, fostering greater inclusion and mutual understanding.

\begin{figure}[t]
  \centering
   \includegraphics[width=0.85\linewidth]{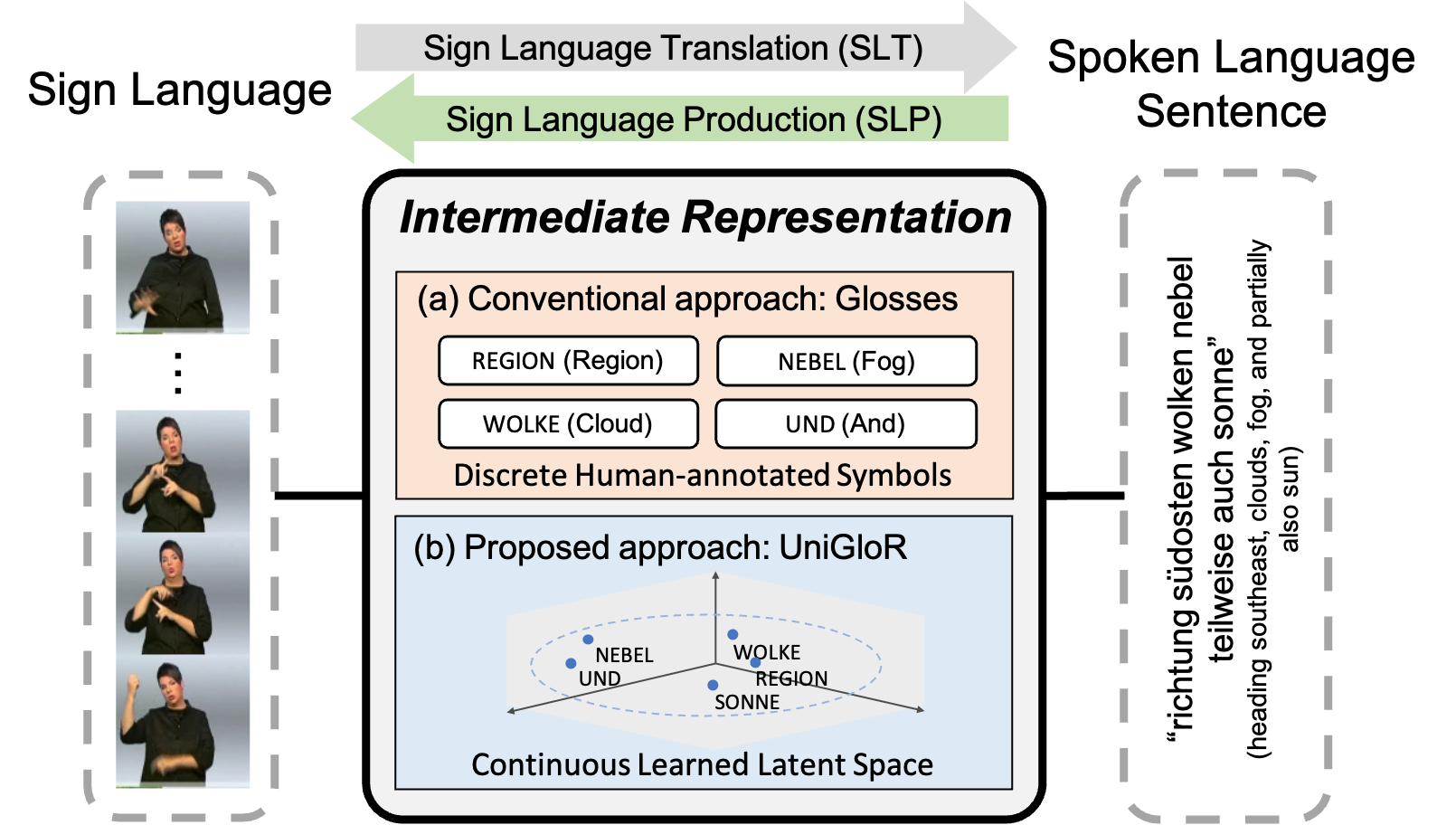}
   \caption{Comparison of the conventional and proposed approaches: (a) incorporating glosses as intermediates, and (b) UniGloR, which uses a learned latent space as intermediates, but not using gloss annotation.}
   \label{fig:gloss_relationship}
   \vspace{-1em}
\end{figure}

In sign language research, significant attention has been paid to Sign Language Translation (SLT)~\cite{camgoz2020sign,voskou2021stochastic,zhou2021spatial,zhou2021improving,yin2021simulslt,jin2022prior,chen2022simple,chen2022two,zhang2023sltunet} and Sign Language Production (SLP)~\cite{stoll2018sign,zelinka2020neural,saunders2020progressive,huang2021towards,tang2022gloss,arkushin2023ham2pose,xie2024g2p}, both of which are essential for the bidirectional translation system.
SLT aims to convert sign videos or keypoint sequences into spoken language sentences, while SLP works in the opposite direction, generating keypoint sequences from spoken language sentences. 
Traditionally, as shown in~\cref{fig:gloss_relationship} (a), both SLT and SLP methods have relied on glosses---written representations of signs using corresponding words---as an intermediate representation, facilitating translation in both directions. 

However, traditional glosses present two major limitations: 
(i) Annotating the glosses is a labor-intensive and time-consuming process that requires specialized expertise in sign language~\cite{li2020tspnet,shi2022open,lin2023gloss}. This significantly hinders the expansion of sign language datasets and slows the progress of machine learning advancements~\cite{bragg2019sign,yin2021including}.
(ii) The glosses oversimplify the rich, dynamic nature of sign language by failing to capture its essential spatio-temporal dynamics~\cite{huerta2022kosign,kim2024signbleu}. 
For example, the sentence ``The cat is sitting on the table'' is typically mapped into ``CAT SIT TABLE'' in a gloss form. 
This simplification omits details, such as the use of classifiers that show the spatial relationship between the cat and the table, the specific handshapes that indicate the position and action of the cat, and the facial expressions that may convey the cat’s state, such as being relaxed. 
Specifically, the spatial arrangement, where the sign for ``TABLE'' is placed in a specific location and ``CAT'' is signed above it to indicate that the cat is sitting on the table, is completely lost in the gloss. 
These limitations show the need for more sophisticated and comprehensive notation systems that can accurately capture the full complexity of sign language.

Motivated by these concerns, we aim to address a key research question: 
\textit{How can we replace traditional glosses in SLT and SLP to enhance efficiency and effectiveness?}
In response, we introduce a novel framework called \textbf{Uni}versal \textbf{Glo}ss-level \textbf{R}epresentation (\textbf{UniGloR}), which automatically captures the spatio-temporal features inherent in sign language, offering a more dynamic and detailed alternative to the use of the glosses. 
As shown in~\cref{fig:gloss_relationship} (b), the core idea is simple: We derive dense spatio-temporal representations from the sign keypoint sequences through Self-Supervised Learning (SSL). These learned representations serve as intermediaries between sign language and spoken language sentences. 
Specifically, our method begins by pretraining a Masked Autoencoder (MAE)~\cite{devlin2018bert,chen2020generative,dosovitskiy2021image,bao2021beit,xie2022simmim,he2022masked} on randomly selected segments from full sign pose sequences. 
This pretraining is conducted using multiple datasets of sign language, ensuring that the model can generalize well across different signs and contexts, making it scalable. 
Additionally, we introduce Adaptive Pose Weights (APW) during pretraining, which are specifically designed to capture the subtle details of signing motion by assigning proportional attention to the movements of each part of the body. 
Once the pretraining is complete, the resulting spatio-temporal representations remain fixed and are utilized as replacements for the glosses in the task-specific mapping processes for SLT and SLP tasks. By using these rich representations instead of the glosses, our method aims to preserve the full complexity of sign language in the subsequent tasks.

Our contributions can be summarized as follows:
\begin{itemize}[itemsep=0.3mm, parsep=1pt]
    \item We introduce UniGloR, the first SSL-based framework designed to replace glosses. Our method derives dense spatio-temporal representations from sign keypoint sequence, enabling its application in both SLT and SLP.
    \item We propose APW, a method designed to effectively capture the subtle nuances of signing motions by assigning proportional attention to each body part. This ensures that the unique contributions of different body parts are accurately represented, enhancing the quality and precision of keypoint-based methods.    
    \item Our experiments in a keypoint-based setting demonstrate that our method either outperforms or matches the performance of previous gloss-free SLT and SLP methods on two widely-used datasets: PHOENIX14T and How2Sign.   
\end{itemize}

\begin{figure*}[t]
\centering
\begin{minipage}[b]{0.47\linewidth}
    \begin{subfigure}{\linewidth}
        \includegraphics[width=\textwidth]{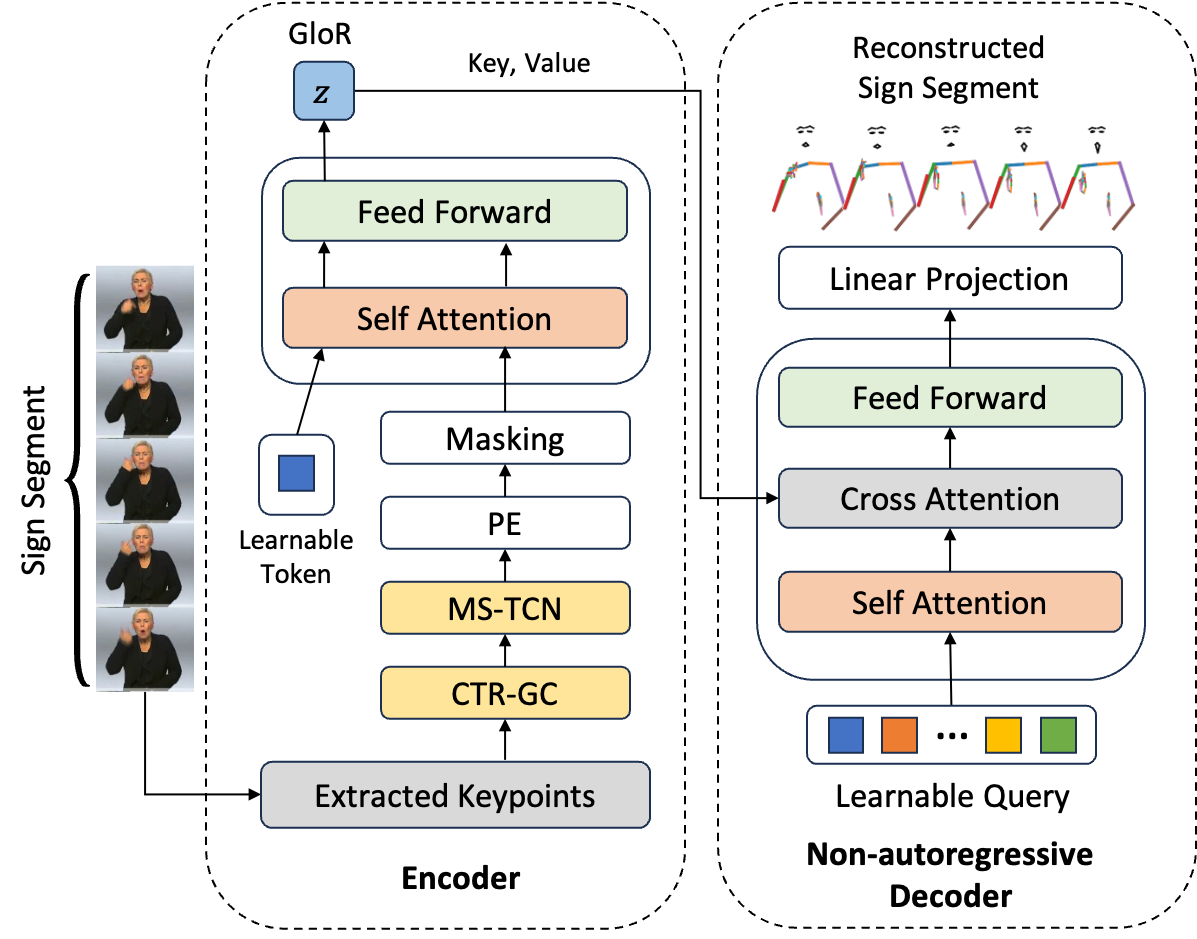}
        \caption{SignMAE}\label{fig:sign_mae}
    \end{subfigure}
\end{minipage}
\hfill
\begin{minipage}[b]{0.50\linewidth}
    \begin{subfigure}{\linewidth}
        \includegraphics[width=\textwidth]{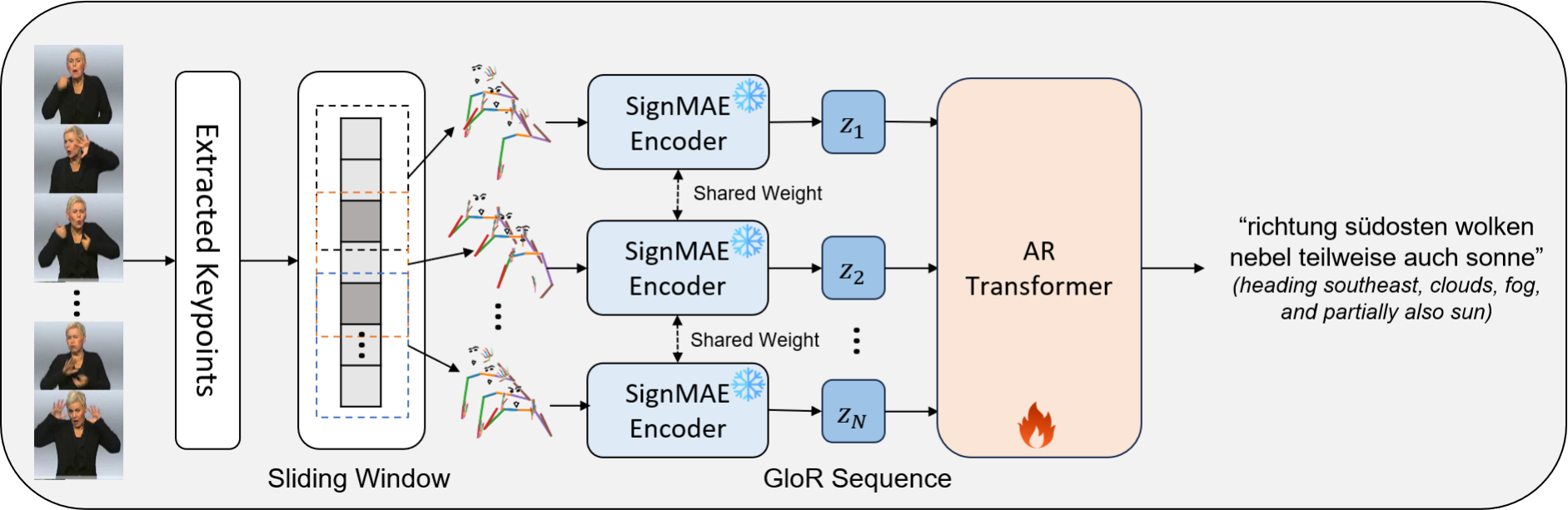}
        \caption{Sign-to-Text Mapping}\label{fig:t2s}
    \end{subfigure}
    \vfill\vspace{0.5em}
    \begin{subfigure}{\linewidth}
        \includegraphics[width=\textwidth]{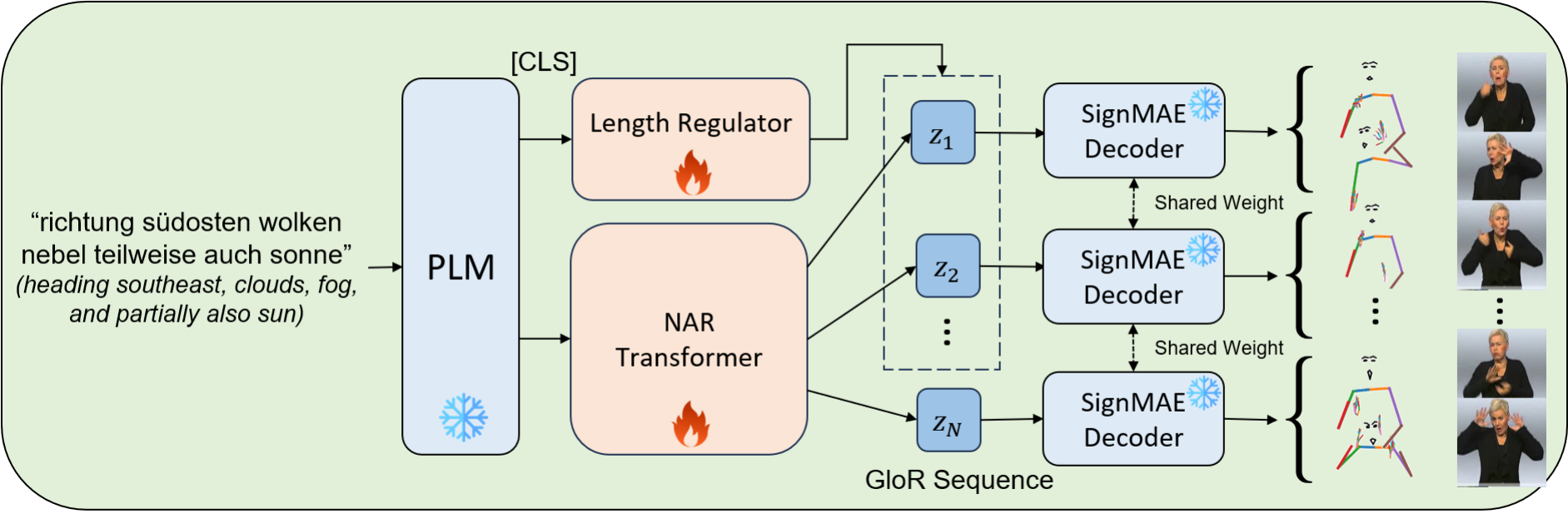}
        \caption{Text-to-Sign Mapping}\label{fig:s2t}
    \end{subfigure}
\end{minipage}
\caption{An overview of the proposed framework. 
(a) \textbf{SignMAE}: We pretrain SignMAE to derive implicit gloss-level representations \(z\) from randomly sampled sign segments. 
(b) \textbf{Sign-to-Text Mapping:} We employ a sliding window approach to fully leverage and enrich the implicit representations, which we then map autoregressively to spoken language sentences.
(c) \textbf{Text-to-Sign Mapping}: A non-autoregressive approach is utilized to map spoken language sentences to gloss-level representations. Following this, a length regulator adjusts the output length. Finally, the frozen SignMAE decoder generates the final sign keypoint sequences.}\label{fig:uniglor_overview} 
\vspace{-1em}
\end{figure*}

\section{Related Work}\label{sec:related}

\subsection{Sign Language Translation}

Sign Language Translation (SLT) converts sign videos or keypoints into spoken language sentences. Glosses have traditionally played an important role in SLT by aligning visual features and identifying semantic boundaries within continuous sign sequences, thereby enabling more accurate translations~\cite{yin2023gloss,wei2023improving}.
For example, Joint-SLT~\cite{camgoz2020sign} incorporates the glosses within a unified framework for both Sign Language Recognition (SLR) and SLT. 
SignBT~\cite{zhou2021improving} constructs a gloss-to-sign dictionary, which serves as the basis for its back-translation process to augment training data. SimulSLT~\cite{yin2021simulslt} employs a boundary predictor to identify gloss boundaries within the sign sequences. 
SLTUNET~\cite{zhang2023sltunet} leverages the glosses in SLT training through a multifaceted approach, integrating tasks such as sign-to-gloss, gloss-to-text, and sign-to-text translation into a unified framework.
While several gloss-free SLT methods~\cite{li2020tspnet,zhao2021conditional,lin2023gloss,fu2023token,wong2024sign2gpt,gong2024llms,rust2024towards,hwang2024efficient} have been proposed, they generally underperform compared to the gloss-based methods. 
By contrast, the present work aims to eliminate the need for hard-annotated glosses by using SSL to derive spatio-temporal representations of sign language at the gloss level, offering an alternative to the glosses.

\subsection{Sign Language Production}

Sign Language Production (SLP) converts spoken language sentences into various forms of sign language, such as keypoints~\cite{stoll2018sign,zelinka2020neural,saunders2020progressive,huang2021towards,tang2022gloss,arkushin2023ham2pose,xie2024g2p}, sign videos~\cite{saunders2022signing,duarte2022sign,cheng2023cico,xie2024sign}, or 3D avatars~\cite{kim2022sign,baltatzis2024neural}. 
NSLS~\cite{zelinka2020neural} was the first to train a network that directly regresses keypoints, while PT~\cite{saunders2020progressive} adopted the Transformer~\cite{vaswani2017attention}, incorporating a counter decoding method to regress keypoints. Subsequent studies improved upon PT by integrating adversarial training~\cite{saunders2020adversarial} and a mixture of motion primitives~\cite{saunders2021mixed}. NAT-EA~\cite{huang2021towards} and NSLP-G~\cite{hwang2021non} introduced non-autoregressive decoding to enable one-shot decoding, thus avoiding error propagation common in autoregressive methods. GEN-OBT~\cite{tang2022gloss} employed an online back-translation method with a Connectionist Temporal Classification (CTC) decoder to semantically align keypoint sequences with their corresponding glosses. 
Recently, G2P-DDM~\cite{xie2024g2p} and SignVQNet~\cite{hwang2024autoregressive} introduced a two-stage learning process using Vector Quantization~\cite{van2017neural}, converting continuous keypoints into discrete representations.
However, most of these methods still rely on the glosses as the primary input source. Additionally, while several methods~\cite{hwang2021non,xie2024g2p,hwang2024autoregressive} use SSL to derive intermediate representations, they are primarily designed for SLP.
By contrast, our approach eliminates the need for the glosses entirely, making it applicable to both SLP and SLT.

\subsection{Self-Supervised Learning}

In the computer vision domain, SSL is a method used to learn generalizable visual representations, often employing various pretext tasks for pretraining~\cite{doersch2015unsupervised,noroozi2016unsupervised,gidaris2018unsupervised}. 
Masked autoencoders (MAE), a type of denoising autoencoder~\cite{vincent2008extracting}, learn representations by reconstructing the original input from corrupted inputs. 
Drawing inspiration from the success of Masked Language Modeling (MLM)~\cite{devlin2018bert}, which follows a similar conceptual approach, MLM has been widely adapted for use in the vision domain.
For example, iGPT~\cite{chen2020generative} processes a sequence of pixels for causal prediction of subsequent pixels. ViT~\cite{dosovitskiy2021image} employs masked token prediction as a self-supervised training step on large-scale image datasets. BEiT~\cite{bao2021beit} proposed predicting discrete token labels for masked patches, requiring an explicit tokenizer to build the token dictionary. SimMIM~\cite{xie2022simmim} and MAE~\cite{he2022masked} focus on directly predicting the pixels of masked patches without any tokenizer design. 
In SLR, SignBERT~\cite{hu2021signbert} uses masking to capture the hierarchical context of hand gestures and their temporal dependencies, while also reconstructing visual tokens derived from hand poses. Similarly, Sign2Vec~\cite{nc2022addressing} utilizes Dense Predictive Coding to learn high-level semantic representations from sign keypoint sequences.
While these previous methods are specifically designed for SLR, limiting their applicability in broader sign language research, our work expands to cover both SLT and SLP, which have fundamentally different directions.

\section{Method}


\subsection{Framework Overview}\label{sec:framework_overview}

Given a sign keypoint sequence \(X = \{x_i\}_{i=1}^T\), where each keypoint \(x_t \in \mathbb{R}^{V \times C}\) represents a frame with \(V\) vertices and a feature dimension \(C\) (e.g., x and y coordinates), our framework is primarily designed to capture spatio-temporal features \(Z\). These features serve as a bridge between the sign sequence and the target translation \(Y = \{y_i\}_{i=1}^U\), where \(U\) is the number of words.
To achieve this, we employ a simple, effective, and scalable MAE to extract these spatio-temporal features. 
Specifically, our method is trained on randomly sampled sign segments from the keypoint sequences, allowing the model to focus on localized dynamics and preserve the subtle movements and transitions within each sign. 
We refer to the learned representations as \textit{implicit gloss-level representations} because accurately segmenting the sign keypoint sequence into explicit gloss-level segments is challenging without the support of pretrained Continuous Sign Language Recognition (CSLR) models~\cite{wei2023improving}. 
Additionally, training on smaller segments rather than entire sequences enhances computational efficiency, making the process more scalable while still maintaining comprehensive and balanced representations across the entire sign keypoint sequence. To further refine these representations, we introduce APW during pretraining, which captures the intricate signing motions within the segments across both spatial and temporal dimensions. 
For SLT and SLP, we perform task-specific mapping to convert the extracted features to text and vice versa, tailored to the specific requirements of each task.
An overview of these processes is provided in~\cref{fig:uniglor_overview}. Detailed explanations of each component are given in the following sections.

\subsection{SignMAE}\label{sec:sign_mae}

As shown in~\cref{fig:sign_mae}, we employ a Transformer-based MAE to extract an implicit gloss-level representation \(z\) from a randomly sampled sign segment. We use CTR-GC~\cite{chen2021channel} and MS-TCN~\cite{liu2020disentangling} for enriched spatio-temporal feature extraction from the sign segment.
CTR-GC dynamically refines the graph topology for each channel, enabling the model to capture intricate spatial relationships between joints, while MS-TCN focuses on capturing temporal dependencies at multiple scales, effectively recognizing both short-term and long-term patterns in the sequences. 
Once the features are extracted, Positional Embeddings (PE)~\cite{vaswani2017attention} are added, followed by random masking applied to the input sequence, analogous to the approaches in BERT~\cite{devlin2018bert} and MAE~\cite{he2022masked}. A learnable token (e.g., \texttt{[GloR]}) is then prepended to the sequence for token pooling~\cite{dosovitskiy2020vit}, mapping the sign segment into a unified latent space and generating the implicit gloss-level representation. Finally, this processed sequence is fed to a Transformer encoder.

To reconstruct the sign segment from the implicit gloss-level representation, we employ a non-autoregressive decoder. This approach enables faster generation and avoids common issues associated with autoregressive models, such as error propagation~\cite{hwang2021non,huang2021towards}. In contrast to NSLP-G, which uses PE for decoder inputs, we introduce learnable queries \(q\), inspired by~\cite{li2023blip}, to enhance flexibility. These queries interact with the self-attention layers and the encoder output through cross-attention layers, serving as both key and value. Finally, the sign segment is reconstructed after passing through a linear projection layer. To further refine the representation, we introduce APW, which is detailed in the next section.

\subsection{Adaptive Pose Weighting}\label{sec:apw}

In the previous section, we discussed the extraction of gloss-level representations from sign segments. While these representations effectively capture the major movements of signs, they often fall short in capturing the subtle nuances and fine-grained motions that are crucial in sign language. This limitation arises from the use of traditional distance-based loss functions, such as L2 loss~\cite{saunders2020progressive,saunders2020adversarial,saunders2021mixed,huang2021towards,hwang2021non} and L1 loss~\cite{tang2022gloss}, commonly applied in pose estimation~\cite{stenum2021applications} and generation tasks~\cite{zhou2019dance}. These losses tend to prioritize large body movements, often overlooking the finer details that are essential for accurate interpretation. To address this, we introduce APW, which adjusts the weights proportionally based on the movement variance of each body part. This method ensures that subtle motions are given appropriate emphasis, preventing them from being overshadowed by more prominent movements, and balancing the model’s focus between broad and detailed motion patterns.

Specifically, the first step involves splitting the keypoints into three distinct categories: body, face, and hands, represented as \(X=[X^{b}; X^{f}; X^{h}]\). We then calculate the spatio-temporal variances for each category across all frames, denoted as \( \sigma^2_{\textit{b}}, \sigma^2_{\textit{f}},\) and \( \sigma^2_{\textit{h}}\) for the body, face, and hands, respectively. For the hands, we measure residual movement \( \sigma^2_{\textit{rh}} \) by calculating the positional difference between each hand and its corresponding wrists, expressed as \( X^{\textit{rh}} = X^{\textit{h}} - X^{\textit{wrist}}\). 
For the face, we normalize the keypoints \(X^{nf}\) by calculating their spatial centroids \(\mu_f\in\mathbb{R}^{T\times1}\) and adjusting the face keypoints accordingly, expressed as \(X^{nf} = X^{\textit{f}} - \mu_{f}\). 
To assign adaptive weights, we use these variances to measure the dynamic importance of each category. The weight for each category is calculated as: 
\begin{equation}
    w^{p}=1-\frac{\sigma^2_{p}+\sigma^2_{b}}{\sum\sigma^2},\quad p\in\{b,nf,rh\},
\end{equation}
where \(p\) indicates the particular part, and \(\sigma^2_b\), the variance of the body part, serves as the baseline for adjusting the weights of other parts. 

Then, the weights for each frame \(t\) are represented as \( W_t = [W_t^b; W_t^{nf}; W_t^{rh}], \) where \(W_t^p=\{w_1^p, w_2^p, \ldots, w_{N(p)}^p\}\), with \(N(p)\) being the number of keypoints in the categories. These weights \(W\) are applied to ensure a consistent emphasis on each part throughout the sequence. With APW, the total loss of SignMAE is defined as:
\begin{equation}\label{eq:l2loss}
    \mathcal{L}_{mae} = \sum_{t=1}^{N}W_t \left \| \psi_t - \hat{\psi}_t \right \|^2_2.
\end{equation} 
where \(\psi\) and \(\hat{\psi}\) denote the ground-truth and generated sign segment, respectively.

\subsection{Task Specific Mapping}\label{sec:task_mapping}

\paragraph{Sign-to-Text Mapping.}
We recall that the goal of SLT is to translate a sign keypoint sequence \(X\) into a spoken language sentence \(Y = \{y_i\}_{i=1}^U\), where \(U\) represents the number of words. 
To fully leverage the implicit gloss-level representations \(Z\), we employ a sliding window approach.
Specifically, as shown in~\cref{fig:s2t}, the sign keypoint sequence is divided into short, overlapping segments. Each segment is then fed into the pretrained SignMAE encoder to extract implicit gloss-level features, represented as \(Z=\{z_i\}_{i=1}^N\), where \(N\) is the number of segments. The number of segments \(N\) is calculated as \(N = \left \lfloor \frac{T-w}{s} \right \rfloor + 1\), where \(T\) is the total number of frames, and \(w\) and \(s\) are the window size and stride, respectively. Finally, we train a Transformer encoder-decoder model using cross-entropy loss, represented as:
\begin{equation}
    \mathcal{L}_\textit{s2t} = -\sum_{i=1}^{U} \sum_{j=1}^V y_{i,j}\log (\hat{y}_{i,j}),
\end{equation}
where \(V\) is the size of the vocabulary and \(\hat{y}\) denotes predicted probability distribution.

\vspace{-1em}
\paragraph{Text-to-Sign Mapping.}
For adopting such representation to SLP, we also use a Transformer encoder-decoder model to map spoken language sentences to the gloss-level representation \(Z\), similar to the sign-to-text mapping. 
However, inspired by NSLP-G~\cite{hwang2021non}, we adopt a non-autoregressive decoder to address the limitations associated with autoregressive methods~\cite{saunders2020progressive,saunders2020adversarial,saunders2021mixed}, particularly when handling continuous representations.
However, NSLP-G cannot adjust the length of the output sequences, which is important in modeling variable lengths in SLP. To address this, we introduce a Length Predictor module, which has a simple architecture with a bottleneck-style fully connected layer and residual connections. Details of the Length Regulator module can be found in~\cref{sec:appen_net_details}.

As shown in~\cref{fig:t2s}, we begin by encoding the spoken language sentence \(Y\) using a Pretrained Language Model (PLM), such as BERT~\cite{devlin2018bert}. Then, the encoded output, excluding the class head token (e.g., \texttt{[CLS]}), is fed into the non-autoregressive Transformer to generate gloss-level representations. We also utilize a fixed set of learnable queries as input to the decoder, similar to SignMAE. The head token, on the other hand, is employed in the Length Regulator module to generate a valid mask, which adjusts the decoder output through element-wise multiplication. The adjusted output is then transformed into an actual sign keypoint sequence \(X\) using the frozen SignMAE decoder.
These modules are jointly trained using two losses: a representation loss and a length loss. The representation loss, which is an L2 loss, measures the difference between the generated representation  and the representation produced by the frozen SignMAE encoder. This is formulated as:
\begin{equation}
    \mathcal{L}_{repr} = \sum_{i=1}^{N} \left \| z_i - \hat{z}_i \right \|^2_2,
\end{equation}
where \(N\) is the number of the gloss-level representations, and \(\hat{z}\) denotes a generated representation. 

The length loss is formulated using Binary Cross Entropy (BCE) loss as:
\begin{equation}
    \mathcal{L}_{len} = -\sum_{i=1}^{N} [l_{i} \cdot \log(p_{i}) + (1 - l_{i}) \cdot \log(1 - p_{i})],
\end{equation}
where \(l_{i}\) is the ground truth binary mask value for the \(i^{th}\) token (1 for valid, 0 for invalid), and \(p_{i}\) is the predicted probability of the \(i^{th}\) token being valid. 
The total loss is the sum of the sequence loss and length loss:
\begin{equation}
    \mathcal{L}_\textit{t2s} = \mathcal{L}_{repr} + \mathcal{L}_{len}.
\end{equation}

\section{Experimental Settings}

\subsection{Dataset and Evaluation Metrics}

\paragraph{Datasets.}
We evaluate our method on two sign language datasets: \textbf{PHOENIX14T}~\cite{camgoz2018neural} and \textbf{How2Sign}~\cite{duarte2021how2sign}. PHOENIX14T is a German Sign Language (DGS) dataset focused on weather forecasts, representing a closed domain with a vocabulary of 3K words and an average of 116 frames. By contrast, How2Sign is a large-scale American Sign Language (ASL) dataset that covers a more open instructional domain, with a vocabulary of 16K words and an average of 173 frames. 
Since PHOENIX14T does not include skeleton keypoints, we followed previous works~\cite{saunders2020progressive,hwang2021non,hwang2024autoregressive} and extracted keypoints using OpenPose~\cite{cao2021realtime} and applied a skeleton correction model~\cite{zelinka2020neural}. 
Detailed statistics for both datasets can be found in~\cref{sec:ds_stats}.

\vspace{-1em}
\paragraph{Metrics.}
We utilized BLEU~\cite{papineni2002bleu} and ROUGE-L~\cite{lin2004automatic} as evaluation metrics for both SLT and SLP. BLEU-n measures translation precision by evaluating n-grams, while ROUGE-L assesses text similarity by calculating the F1 score based on the longest common subsequences between predicted and reference texts. To measure BLEU in SLP, we employed Back-Translation (BT)~\cite{saunders2020progressive}, which involves translating the generated sign keypoint sequences back into spoken language and comparing them to the original text. We trained Joint-SLT~\cite{camgoz2020sign} on PHOENIX14T and How2Sign, following previous works~\cite{saunders2020progressive,hwang2021non,hwang2024autoregressive,huang2021towards,tang2022gloss}.
Additionally, we evaluated the Mean Per Joint Position Error (MPJPE), commonly used in human motion prediction tasks~\cite{cai2020learning,li2022skeleton,guo2023back}.

\subsection{Implementation Details}

\paragraph{Keypoint Processing.}
We employed an abstract representation of keypoints that emphasizes essential elements while omitting less relevant features, such as facial lines and the nose, to focus on the critical aspects of sign language interpretation. This representation includes a total of 73 keypoints: 19 for the face, 23 for each hand, and 8 for the upper body. The keypoints undergo several preprocessing steps, including centering, normalization, and removal of noisy frames. Further details of these processes are provided in~\cref{sec:appen_preprocessing}.

\vspace{-1em}
\paragraph{Training.}
SignMAE is trained on both PHOENIX14T and How2Sign, with the sign segment size set to 15 frames. This choice aligns with previous findings~\cite{wilbur2009effects}, which reveal that approximately 15 frames are ideal for capturing a single sign in videos recorded at 30 frames per second (FPS). 
For sign-to-text mapping, we adopted a stride of 7 to achieve denser and enriched gloss-level representations. 
In contrast, text-to-sign mapping employed a stride of 15, consistent with the segment size used in SignMAE, to ensure precise reconstruction of gloss-level representations into sign keypoint sequences via the pretrained SignMAE decoder. 
The selection of these stride sizes reflects the distinct requirements of each task. 
Further details about each network are provided in~\cref{sec:appen_net_details}.

\section{Main Results}

\begin{table}[t]
    \scriptsize
    \centering
    \begin{subtable}{\linewidth}
        \resizebox{\linewidth}{!}{%
            \begin{tabular}{lcccccc} \toprule
                \textbf{Methods}& \textbf{Gloss}& \textbf{BLEU-1}& \textbf{BLEU-2}& \textbf{BLEU-3}& \textbf{BLEU-4}& \textbf{ROUGE} \\ \midrule
                Joint-SLT~\cite{camgoz2020sign}& \xmark& 21.31& 14.17& 10.05& 8.12& 22.56 \\
                ConSLT~\cite{fu2023token}& \xmark& 21.09 & 13.93 & 10.48 & 8.49 &21.81 \\
                Joint-SLT~\cite{camgoz2020sign}\(\dagger\)& $\checkmark$& \underline{24.12}& \underline{16.77}& \underline{12.80}& \underline{10.58}& \underline{27.70} \\
                \noalign{\vskip 0.3ex}\cdashline{1-7}\noalign{\vskip 0.7ex}
                Ours& \xmark& \textbf{30.24}& \textbf{20.96}& \textbf{15.99}& \textbf{12.86}& \textbf{29.65} \\ \bottomrule
            \end{tabular}%
        }
        \caption{Translation Results on PHOENIX14T}\label{tab:slt_res_p14t}
    \end{subtable}
    \vfill \vspace{1em}
    \begin{subtable}{\linewidth}
        \resizebox{\linewidth}{!}{%
            \begin{tabular}{lcccccc} \toprule
                \textbf{Methods}& \textbf{Gloss}& \textbf{BLEU-1}& \textbf{BLEU-2}& \textbf{BLEU-3}& \textbf{BLEU-4}& \textbf{ROUGE} \\ \midrule
                Joint-SLT~\cite{camgoz2020sign}& \xmark& 11.11& 4.41& 1.98& 0.78& 10.61 \\
                GloFE-VN~\cite{lin2023gloss}& \(\triangle\)& \underline{14.94} & \underline{7.27} &\textbf{3.93} &\textbf{2.24} & \underline{12.61} \\
                \noalign{\vskip 0.3ex}\cdashline{1-7}\noalign{\vskip 0.7ex}
                Ours& \xmark& \textbf{18.52}& \textbf{7.88}& \underline{3.83}& \underline{2.22}& \textbf{12.98} \\ \bottomrule
            \end{tabular}%
        }
        \caption{Translation Results on How2Sign}\label{tab:slt_res_h2s}
    \end{subtable}
\caption{Translation results on PHOENIX14T and How2Sign. \(\dagger\) indicates results reproduced by~\cite{huang2021towards}, while \(\triangle\) marks where the visual encoder was pretrained with glosses. The best results are shown in \textbf{bold}, with the second-best \underline{underlined}.}\label{tab:slt_results}
\vspace{-1em}
\end{table}

\paragraph{Translation Results.}

We evaluated our method against three baseline methods: Joint-SLT~\cite{camgoz2020sign}, ConSLT~\cite{fu2023token}, and GloFE-VN~\cite{lin2023gloss}. Since Joint-SLT and ConSLT were originally designed for sign video inputs, we adapted both models to work with sign keypoint inputs and reproduced their results. For consistency, we also report the results reproduced by~\cite{huang2021towards}. 
As shown in~\cref{tab:slt_res_p14t}, our method outperforms Joint-SLT on PHOENIX14T, with improvements of 2.28 in BLEU-4 and 1.95 in ROUGE. 
On How2Sign (see~\cref{tab:slt_res_h2s}), the improvement is even more pronounced, though our results are competitive against rather than superior to GloFE-VN, which incorporates the glosses during pretraining for its visual encoder~\cite{hwang2024efficient}. 
Specifically, our method shows an improvement of 0.37 in ROUGE, but a slight 0.02 decrease in BLEU-4 compared to GloFE-VN. 
While we do not claim state-of-the-art performance, our focus is on introducing a novel gloss-level representation as a robust alternative to traditional glosses, demonstrating competitive, and in some cases, superior performance within a keypoint-based setting.

\begin{table}[t]
    \centering
    \scriptsize
    \begin{subtable}{\linewidth}
        \resizebox{\linewidth}{!}{%
            \begin{tabular}{lccccccc} \toprule
                \textbf{Method}& \textbf{Gloss}& \textbf{BLEU-1}& \textbf{BLEU-2}& \textbf{BLEU-3}& \textbf{BLEU-4}& \textbf{ROUGE}& \textbf{MPJPE\(\downarrow\)} \\ \midrule
                PT~\cite{saunders2020progressive}& \xmark&12.07& 5.94& 4.04& 3.08& 11.06& 0.602 \\
                NSLP-G~\cite{hwang2021non}& \xmark& \underline{19.10}& \underline{12.10}& \underline{8.87}& \underline{6.97}& \underline{18.62}& 0.335 \\ 
                PT~\cite{saunders2020progressive}& \(\checkmark\) &14.84 &8.00 &5.49 &4.19 &12.83& 0.590 \\
                NSLP-G~\cite{hwang2021non}& \(\checkmark\)& 18.25& 11.06& 7.86& 6.14& 17.71& \underline{0.332} \\
                \noalign{\vskip 0.3ex}\cdashline{1-8}\noalign{\vskip 0.7ex}
                Ours& \xmark& \textbf{29.60}& \textbf{21.25}& \textbf{16.48}& \textbf{13.44}& \textbf{31.38}& \textbf{0.183} \\ \bottomrule
            \end{tabular}
        }
        \caption{Production Results on PHOENIX14T}
    \end{subtable}
    \vfill \vspace{1em}
    \begin{subtable}{\linewidth}
        \resizebox{\linewidth}{!}{%
            \begin{tabular}{lccccccc} \toprule
                \textbf{Method}& \textbf{Gloss}& \textbf{BLEU-1}& \textbf{BLEU-2}& \textbf{BLEU-3}& \textbf{BLEU-4}& \textbf{ROUGE}& \textbf{MPJPE\(\downarrow\)} \\ \midrule
                PT~\cite{saunders2020progressive}& \xmark& \underline{11.88} & \underline{4.50}& \underline{1.75}& 0.42& 9.76& 0.466 \\
                NSLP-G~\cite{hwang2021non}& \xmark& 9.73& 3.67& 1.53& \underline{0.48}& \underline{10.18}& \underline{0.427} \\ 
                \noalign{\vskip 0.3ex}\cdashline{1-8}\noalign{\vskip 0.7ex}
                Ours& \xmark& \textbf{12.69}& \textbf{5.12}& \textbf{2.36}& \textbf{1.06}& \textbf{10.93}& \textbf{0.169} \\ \bottomrule
            \end{tabular}
        }
        \caption{Production Results on How2Sign}
    \end{subtable}
    \caption{Production results on PHOENIX14T and How2Sign.}
    \label{tab:slp_results}
\end{table}
\vspace{-1em}
\paragraph{Production Results.}
We compare our method against two baselines: PT~\cite{saunders2020progressive} and NSLP-G~\cite{hwang2021non}. We adapted PT by excluding the use of additional ground-truth data, in line with recommendations from previous studies~\cite{huang2021towards,tang2022gloss,hwang2024autoregressive}. 
As shown in~\cref{tab:slp_results}, our method consistently outperforms these baselines across all evaluation metrics on the datasets. 
Notably, on PHOENIX14T, our approach achieves significant improvements, with BLEU-4 and MPJPE that are 6.47 and 0.149 points better than those of NSLP-G, respectively. On How2Sign, our method also surpasses the baselines, with gains of 0.58 in BLEU-4 and a reduction of 0.258 in MPJPE. 
These results demonstrate the effectiveness of our method in SLP, not only improving accuracy but also capturing the temporal and contextual subtleties of sign language, while still maintaining its applicability to SLT.

\begin{table}[t]
\centering
\scriptsize
\resizebox{0.85\linewidth}{!}{%
    \begin{tabular}{lccc}
    \toprule
    \textbf{Methods} & \textbf{Top-1}& \textbf{Top-5}& \textbf{Top-10} \\ \midrule
    GRU + Normalization~\cite{kim2022openpose}& 90.46 
    & -&- \\
    Multi Transformer Encoder~\cite{roh2023transformer}& \underline{91.01}  & -&- \\ \noalign{\vskip 0.3ex}\cdashline{1-4}\noalign{\vskip 0.7ex}
    Ours& \textbf{95.03}& \textbf{97.82}&\textbf{98.00}\\ \bottomrule
    \end{tabular}%
}
\caption{Recognition results on KSL-Guide-Word.}\label{tab:slr_results}
\vspace{-1em}
\end{table}
\vspace{-1em}
\paragraph{Out-of-domain Results.}
We extended our method to SLR to evaluate its performance on out-of-domain tasks with unseen data, aiming to assess the robustness of the gloss-level representations. For this evaluation, we utilized the KSL-Guide-Word dataset~\cite{ham2021ksl}, which focuses on the transportation and direction domain and comprises 3K words.
For the task-specific mapping, we employed a Transformer encoder configured identically to the SignMAE encoder, followed by a single linear projection layer that maps the encoded features to class probabilities.
As shown in~\cref{tab:slr_results}, our method significantly outperforms baseline methods, achieving a 4.02 improvement in Top-1 accuracy, representing a 4.42\% performance gain.

\section{Ablation Study}
To further evaluate our proposed method, we conducted extensive ablation experiments on PHOENIX14T, the most widely used dataset in sign language research. In particular, we examined the key components of SignMAE, APW and stride size.

\begin{table}[t]
\scriptsize
\centering
\begin{subtable}[b]{0.49\columnwidth}
    \resizebox{\columnwidth}{!}{%
    \begin{tabular}{lcc} \toprule
    \textbf{Decoding Type}& \textbf{BLEU-4}& \textbf{ROUGE} \\ \midrule
     AR &12.02& 28.99 \\
     NAR (Ours)& \textbf{12.86}& \textbf{29.65} \\ \bottomrule
    \end{tabular}%
    }
    \caption{Decoding Type}
    \label{tab:abl_dec}
\end{subtable}
\hfill
\begin{subtable}[b]{0.5\columnwidth}
    \resizebox{\columnwidth}{!}{%
    \begin{tabular}{lcc} \toprule
    \textbf{Query Type}& \textbf{BLEU-4}& \textbf{ROUGE}\\ \midrule
     PE~\cite{hwang2021non}& 11.32& 28.09 \\
     Learnable (Ours)& \textbf{12.86}& \textbf{29.65} \\ \bottomrule
    \end{tabular}%
    }
    \caption{Query Type}
    \label{tab:abl_query_type}
\end{subtable}
\vfill
\vspace{0.5em}
\begin{subtable}[b]{0.49\linewidth}
    \resizebox{\columnwidth}{!}{%
    \begin{tabular}{lcc} \toprule
    \textbf{Window Size} &\textbf{BLEU-4}& \textbf{ROUGE} \\ \midrule
     N=32& 10.59& 26.87 \\
     N=15 (Ours)& \textbf{12.86}& \textbf{29.65} \\ \bottomrule
    \end{tabular}%
    }
    \caption{Window Size}\label{tab:abl_window}
\end{subtable}
\begin{subtable}[b]{0.47\linewidth}
    \resizebox{\textwidth}{!}{%
    \begin{tabular}{lccc}
    \toprule
    \textbf{Dataset} &\textbf{BLEU-4}& \textbf{ROUGE} \\ \midrule
     P14T &10.23 &26.31 \\
     P14T+H2S& \textbf{12.86}& \textbf{29.65} \\ \bottomrule
    \end{tabular}%
    }
    \caption{Dataset Used}\label{tab:abl_ds}
\end{subtable}
\caption{Ablation study for main components of SignMAE. P14T and H2S represent PHOENIX14T and How2Sign, respectively.}\label{tab:ablation}
\end{table}
\vspace{-1em}
\paragraph{Impact of Main Components.}
We conducted an analysis to determine how different components of SignMAE influence SLT performance, focusing on (i) decoding strategies, (ii) input query types, (iii) window size, and (iv) the dataset utilized.
Our findings indicate that non-autoregressive decoding slightly outperforms autoregressive decoding, suggesting that autoregression may not be essential, as shown in~\cref{tab:abl_dec}. 
Additionally, the use of a learnable query marginally surpasses PE used in~\cite{hwang2021non}, as shown in~\cref{tab:abl_query_type}.
Increasing the window size reduced focus on key features, supporting our choice based on~\cite{wilbur2009effects}, as shown in~\cref{tab:abl_window}. 
Last, our analysis shows that scaling the dataset produced substantial performance improvements, specifically an increase of 2.63 in BLEU-4 and 3.34 in ROUGE scores, as demonstrated in~\cref{tab:abl_ds}.

\begin{table}[t]
\scriptsize
\centering
\begin{subtable}{\linewidth}
    \resizebox{\linewidth}{!}{%
    \begin{tabular}{lccccc} \toprule
        \textbf{Models}& \textbf{BLEU-1}& \textbf{BLEU-2}& \textbf{BLEU-3}& \textbf{BLEU-4}&  \textbf{ROUGE} \\ \midrule
         Joint-SLT~\cite{camgoz2020sign}& 21.31 & 14.17 & 10.05 & 8.12 & 22.56
        \\
        \quad w/ FPW& 23.45 & 15.16 & 10.95 & 8.52 & 
        22.78
        \\
        \quad w/ APW& \textbf{24.22} & \textbf{15.92} & \textbf{11.61} & \textbf{9.11} & \textbf{23.47}
        \\ \midrule
         
        ConSLT~\cite{fu2023token}& 21.09 & 13.93 & 10.48 & 8.49& 21.81 \\
        \quad w/ FPW& 21.16 & 14.09 & 10.53 & 8.50 & 21.50
        \\
        \quad w/ APW& \textbf{22.16}& \textbf{14.91}& \textbf{11.15}& \textbf{9.00}& \textbf{22.18}\\ \midrule

        Ours&         28.73& 19.51 & 14.25 & 11.80 &27.75\\
        \quad w/ FPW& 29.16& 20.04 & 15.07 & 12.04 &28.93\\
        \quad w/ APW& \textbf{30.24}& \textbf{20.96}& \textbf{15.99}& \textbf{12.86}& \textbf{29.65} \\ \bottomrule
    \end{tabular}%
    }
    \caption{Translation Performance}\label{tab:apw-slt}
\end{subtable}
\vfill \vspace{0.5em}
\begin{subtable}{\linewidth}
    \resizebox{\linewidth}{!}{%
    \begin{tabular}{lcccccc} \toprule
        \textbf{Models}& \textbf{BLEU-1}& \textbf{BLEU-2}& \textbf{BLEU-3}& \textbf{BLEU-4}& \textbf{ROUGE} \\ \midrule
         
         PT~\cite{saunders2020progressive}& 12.07& 5.94& 4.04& 3.08& 11.06 \\
         \quad w/ FPW& 13.45& 6.88& 4.65& 3.47& 12.03 \\
         \quad w/ APW& \textbf{13.52}& \textbf{7.12}& \textbf{5.03}& \textbf{3.99}& \textbf{12.14} \\ \midrule
        
        NSLP-G~\cite{hwang2021non}& 19.10& 12.10& 8.87& 6.97& 18.62 \\
        \quad w/ FPW& 18.98& 12.15& 8.87& 6.96& 18.47 \\
        \quad w/ APW& \textbf{19.80}& \textbf{12.56}& \textbf{9.06}& \textbf{7.11}& \textbf{19.37} \\ \midrule
        
        Ours& 25.03& 17.59& 13.71& 11.23& 26.95 \\
        \quad w/ FPW& 27.62& 19.64& 15.12& 12.34& 29.64 \\
        \quad w/ APW& \textbf{29.60}& \textbf{21.25}& \textbf{16.48}& \textbf{13.44}& \textbf{31.38} \\ \bottomrule
    \end{tabular}%
    }
    \caption{Production Performance}\label{tab:apw-slp}
\end{subtable}
\caption{An analysis of the impact of APW. FPW denotes Fixed Pose Weights which are derived from dataset.}\label{tab:abl_apw}
\vspace{-1em}
\end{table}

\vspace{-1em}
\paragraph{Impact of APW.}
We extended our evaluation of APW beyond our methods to include previous methods in both SLT and SLP. We established three distinct scenarios: (i) one without weighting (\(W=\mathbf{1}\)), (ii) one with Fixed Pose Weights (FPW) using static weights for all training samples, and (iii) one with APW. For (ii), we determined the weights based on overall dataset statistics, assigning a weight of 1.0 to the body, 1.17 to the face, and 1.18 to the hands. 
As shown in~\cref{tab:abl_apw}, incorporating APW consistently improves both SLT and SLP methods on all metrics. 
In SLT, our method achieves a 1.06 improvement in BLEU-4, showing an 8.98\% performance increase. Similarly, Joint-SLT benefits from the APW approach with a 0.99 improvement in BLEU-4, resulting a 12.19\% gain. ConSLT also shows a 0.51 increase in BLEU-4, resulting in a 6.01\% performance improvement.
In SLP, the impact of APW is particularly significant. Our method achieves a significant improvement of 2.21 in BLEU-4, a 19.68\% increase. PT experiences an even larger boost, with a 0.91 improvement in BLEU-4, resulting 29.5\% increase. NSLP-G, while showing more modest gains, still improves by 0.14 in BLEU-4, resulting in a 2.01\% performance gain. 
These results demonstrate the effectiveness of APW in enhancing model accuracy and robustness across a range of tasks.

\begin{figure}[t]
    \centering
    \includegraphics[trim=0cm 0cm 0cm 0cm,clip=true,width=\linewidth]{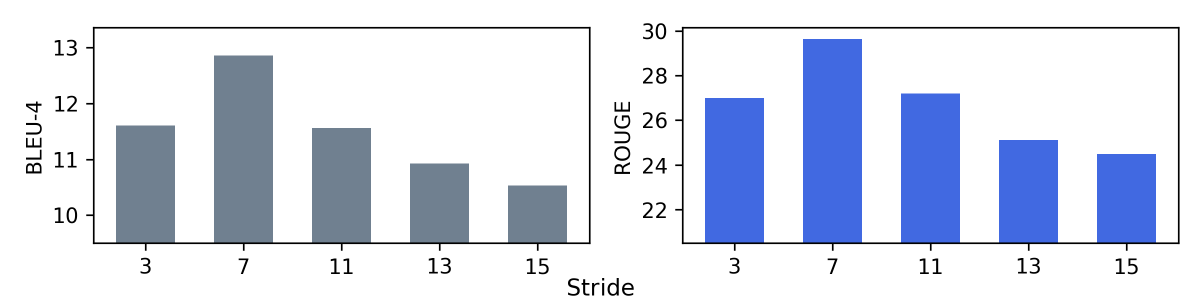}
    \caption{SLT performance curve depends on stride.}\label{fig:abl_window}
\end{figure}

\vspace{-1em}
\paragraph{Impact of Stride.}
Finally, we explored the impact of varying the stride of the sliding window approach in SLT. 
As shown in~\cref{fig:abl_window}, performance improves with increasing stride size, reaching its peak at a stride of 7. 
This stride effectively balances contextual information capture, enhancing overall translation accuracy. 
However, beyond this point, performance begins to decline, likely due to reduced overlap between windows. This reduction leads to coarser representations, causing the model to lose granularity in the spatio-temporal features it captures.

\section{Qualitative Results.}
\begin{table}[t]
    \centering
    \scriptsize
    \resizebox{\linewidth}{!}{%
    \begin{tabular}{rl} \toprule
        Ground Truth:&  und nun die wettervorhersage für morgen donnerstag den sechsundzwanzigsten november.\\
        &(\textit{And now the weather forecast for tomorrow, Thursday the twenty-sixth of November.})\\
        Joint-SLT~\cite{camgoz2020sign}:&  \textcolor{blue}{und nun die wettervorhersage für morgen} \textcolor{red}{samstag} \textcolor{blue}{den} \textcolor{red}{fünfundzwanzigsten dezember}. \\
        &(\textit{And now the weather forecast for tomorrow, Saturday the twenty-fifth of December.})\\  
        ConSLT~\cite{fu2023token}:&  \textcolor{blue}{und nun die wettervorhersage für morgen} \textcolor{red}{freitag} \textcolor{blue}{den} \textcolor{red}{siebenundzwanzigsten mai}.\\
        &(\textit{And now the weather forecast for tomorrow, Friday the twenty-seventh of May.})\\  
        Ours:&  \textcolor{blue}{und nun die wettervorhersage für morgen donnerstag den sechsundzwanzigsten november.}\\
        &(\textit{And now the weather forecast for tomorrow, Thursday the twenty-sixth of November.})\\ \bottomrule
    \end{tabular}
    }
    \caption{Translation results on the testset compared to Joint-SLT and ConSLT on PHOENIX14T. Correctly translated 1-grams are highlighted in \textcolor{blue}{blue}, while incorrect translations are marked in \textcolor{red}{red}.}
    \label{tab:slt_qual_res}
    \vspace{-1em}
\end{table}
\paragraph{Translation Results.}
\cref{tab:slt_qual_res} presents a translation example from PHOENIX14T, comparing our method with the reproduced versions of Joint-SLT and ConSLT under the keypoint setting. Our method demonstrates a more accurate translation, successfully capturing key semantic elements such as ``donnerstag (Thursday)'' and ``november (November)'', where the other baselines fail to do so. Additional results can be found in~\cref{sec:appen_more_trans_res}.

\begin{figure}[t]
    \centering
    \scriptsize
    \centering
    \includegraphics[trim=0cm 0cm 0cm 0cm,clip=true,width=\linewidth]{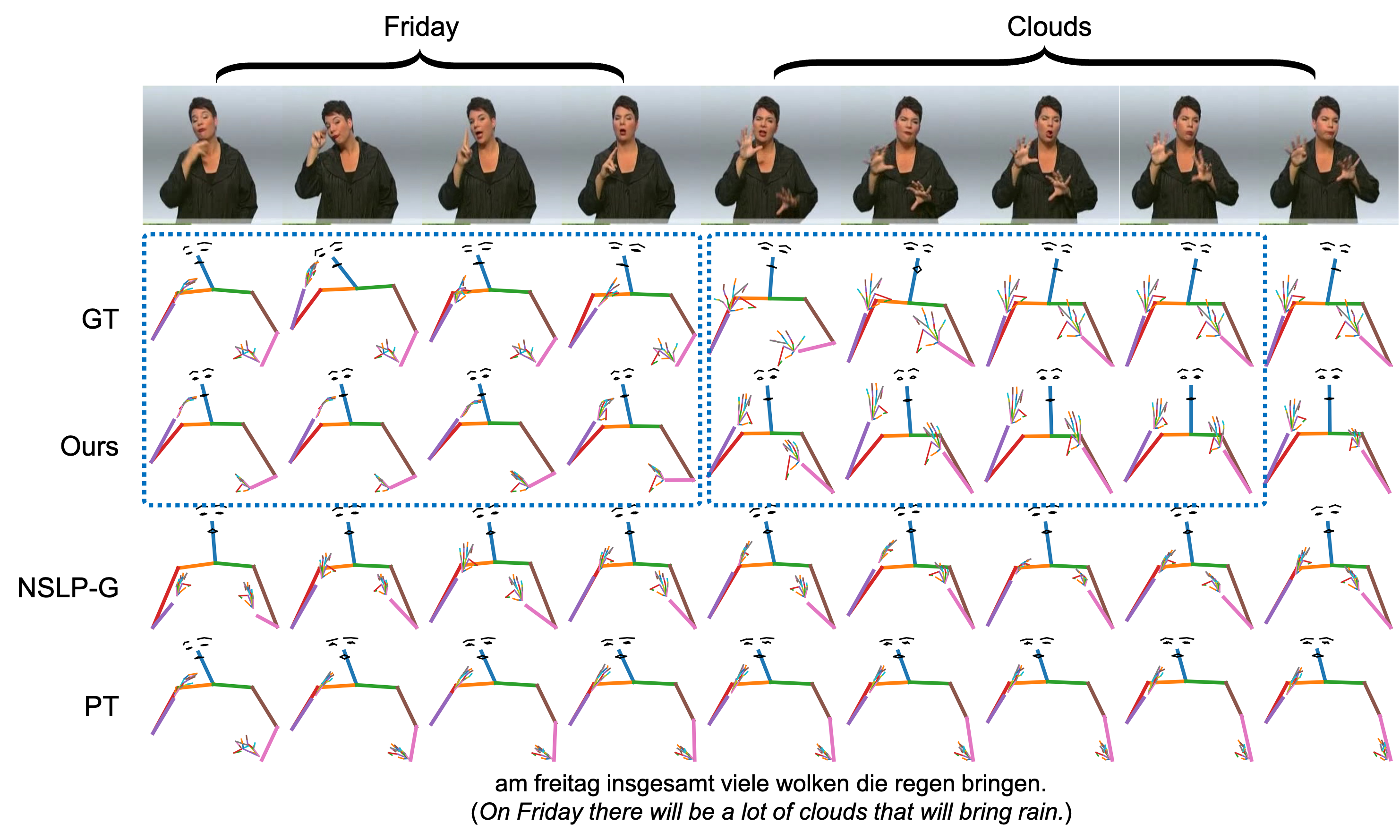}\label{fig:slp-qual}
    \caption{A visual comparison of our method against baselines, including PT and NSLP-G, on PHOENIX14T. We uniformly selected frames. The dashed boxes highlight where our method produce more realistic and accurate sign keypoint sequences. GT refers to the Ground-Truth keypoints.}\label{fig:slp_qual_res}
\end{figure}

\vspace{-1em}
\paragraph{Production Results.}
\cref{fig:slp_qual_res} visually compares the sign keypoint sequences generated by our method against baseline models, including PT and NSLP-G, on PHOENIX14T. The results show that our method consistently produces more accurate and precise representations of signs, such as ``Friday'' and ``Clouds'', demonstrating enhanced performance in both precision and clarity compared to the baselines. Additional results can be found in~\cref{sec:appen_more_prod_res}.

\begin{figure}[t]
    \centering
    \includegraphics[trim=2cm 1.3cm 0.5cm 0.5cm,clip=true, width=0.8\linewidth]{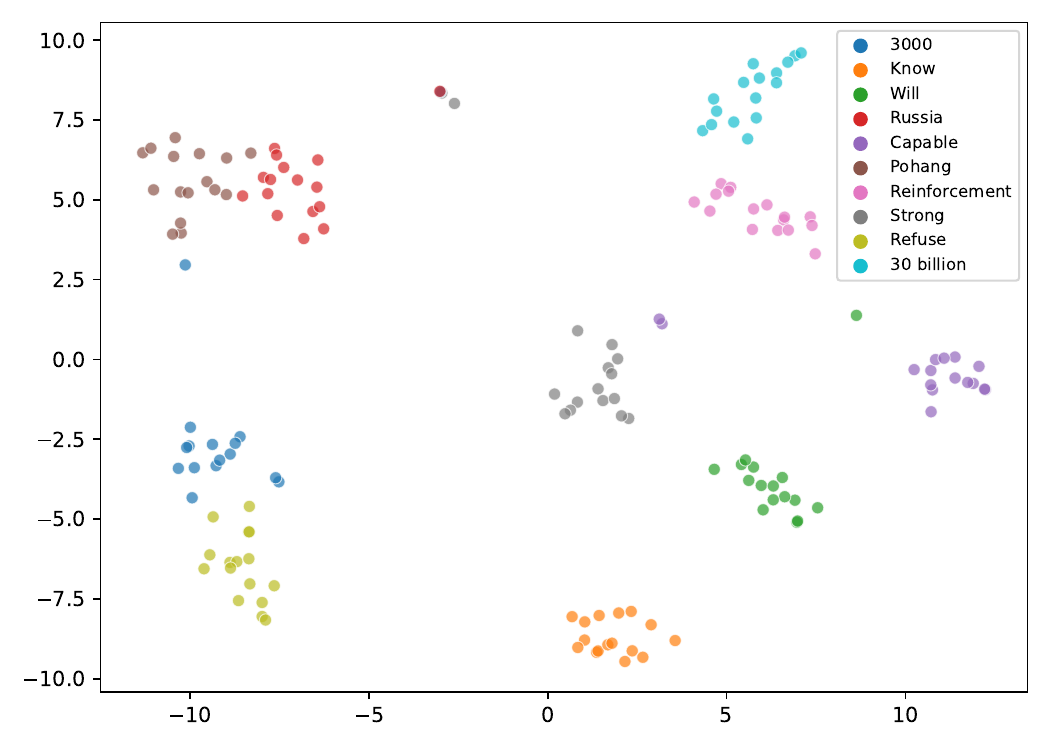}
    \caption{Visualization of gloss-level representations on the unseen KSL-Guide-Word dataset in a zero-shot setting. Ten of the most frequent glosses were selected for this analysis. The 2D plot was generated using T-SNE~\cite{van2008visualizing}.}\label{fig:glor_space}
    \vspace{-1em}
\end{figure}

\vspace{-1em}
\paragraph{Visualization of Learned Representations.}
We present a visualization of gloss-level representations on the KSL-Guide-Word dataset in a zero-shot setting to evaluate the effectiveness of our method in generating distinct representations. Note that PHOENIX14T and How2Sign do not provide gloss-level annotations, making KSL-Guide-Word particularly valuable for this analysis. 
As shown in~\cref{fig:glor_space}, our method produces distinct, well-separated clusters, with each cluster corresponding to a specific gloss, such as ``Know'', ``Will'', ``Russia'', and ``Strong''.

\section{Limitations and Future Work}
In this work, we focus exclusively on using extracted keypoints from the sign videos to minimize the high computational cost associated with processing full RGB data. 
While this keypoint-based approach is computationally efficient and broadly applicable, it comes with a trade-off: some finer details are inevitably lost, which may result in lower performance in SLT compared to methods that utilize the sign video as input. 
Furthermore, in SLP, most methods~\cite{saunders2021anonysign,saunders2022signing} follow a two-step process---first generate sign keypoints from text input, then synthesize realistic images frame-by-frame to produce the final sign video. 
Our decision to utilize keypoints reflects a deliberate compromise intended to address both SLT and SLP effectively.

Currently, our method has been evaluated with a limited number of baselines, primarily due to the restricted modality we used. For example, there are various different modalities, such as sign videos, keypoints, or SMPL-X, making it challenging to reproduce their results under our specific settings. 
Despite these limitations, we have successfully reproduced several popular models, including Joint-SLT, ConSLT, PT, and NSLP-G, to validate our approach.

Scaling datasets has consistently led to performance improvements, as seen with larger sign language datasets, such as Youtube-ASL~\cite{uthus2024youtube}. 
Future work will involve expanding the dataset size to explore the full potential of our method and to assess its scalability and performance across more extensive and diverse datasets.


\section{Conclusion}

In this paper, we introduced UniGloR, a novel framework designed to replace the glosses used in SLT and SLP. We note that the current glosses are not only expensive to annotate but also fall short in fully capturing the complexities of sign language. 
UniGloR addresses these limitations by utilizing SSL to extract rich spatio-temporal representations inherent in sign language.
Furthermore, we propose APW, which captures the subtle signing motions by assigning proportional attention to each body part, thereby enhancing the quality of the representations. 
Our experimental results demonstrate that UniGloR achieves superior and competitive performance compared to existing SLT and SLP methods on PHOENIX14T and How2Sign.

\section*{Acknowledgment}
This work was supported by the the Institute for Information and communications Technology Promotion (IITP, Project No. 2022-0-00010) funded by MSIT, Korea, and the Industrial Fundamental Technology Development Program (No. 20023495) funded by MOTIE, Korea.


{\small
\bibliographystyle{ieee_fullname}
\bibliography{egbib}
}

\clearpage

\appendix
\section*{Appendix}

We begin by providing a detailed analysis of the sign language datasets used in this work in~\cref{sec:ds_stats}. In~\cref{sec:appen_more_imple}, we discuss additional implementation details, including keypoint preprocessing and the network architecture. Further qualitative results are presented in~\cref{sec:appen_more_res}. Lastly, we address the potential negative societal impacts of our work in~\cref{sec:social_impact}.

\section{Statistics of Sign Language Datasets}\label{sec:ds_stats}

We provide details on two sign language datasets used in this work: PHOENIX14T and How2Sign, as summarized in~\cref{tab:dataset_stats}. 
PHOENIX14T is a German Sign Language (DGS) dataset focused on the specific domain of weather forecasting. It features a relatively small vocabulary of 3K words and concise video clips averaging 116 frames in length. The dataset includes 7,096 training samples, 519 validation samples, and 642 test samples, all with gloss annotations. Because it is tailored for domain-specific tasks, PHOENIX14T offers clear and repetitive patterns, making it ideal for translation and recognition tasks within weather-related contexts.
On the other hand, How2Sign is an American Sign Language (ASL) dataset within the instructional domain. This dataset is significantly larger and more diverse, with a vocabulary of 16K words and an average video length of 173 frames. It contains 31,128 training samples, 1,741 validation samples, and 2,322 test samples, though it lacks gloss annotations. The broader and more complex nature of How2Sign makes it well-suited for general sign language processing tasks, particularly those that require an understanding of diverse and intricate sign sequences.

\begin{table*}[t]
\centering
\scriptsize
\resizebox{0.8\linewidth}{!}{%
    \begin{tabular}{lcccccc} \toprule 
     Dataset &Lang &Vocab &Train / Valid / Test &Avg. No.F& Gloss& Domain \\ \midrule 
     PHOENIX14T~\cite{camgoz2018neural}&   DGS&3K& 7,096 / 519 / 642&  116& $\checkmark$& Weather Forecast \\  
     How2Sign~\cite{duarte2021how2sign} &   ASL&16K& 31,128 / 1,741 / 2,322&  173& \xmark& Instructional \\ \bottomrule
    \end{tabular}
    }
\caption{Statistics of Sign Language Datasets. Avg. No.F means average number of frames.} 
\label{tab:dataset_stats}
\end{table*}


\section{More Implementation Details}\label{sec:appen_more_imple}

\subsection{Keypoint Preprocessing}\label{sec:appen_preprocessing}

In preprocessing the keypoints, we first center and normalize them based on the shoulder joint, ensuring that the length of the shoulder is normalized to 1, following~\cite{yoon2019robots}. However, off-the-shelf extraction models, such as OpenPose~\cite{cao2021realtime}, do not always yield consistently high-quality keypoint data. To address this issue, we implemented an additional step to filter out noisy frames—specifically, those with missing or misplaced joints—to ensure data quality and consistency~\cite{Yang2014Effective,Liu2017Skeleton-Based}. A frame is considered noisy if the joint distance between consecutive frames exceeds a certain threshold. To detect such frames, we first calculate the differences between consecutive frames as:
\begin{equation}
    X_{\text{diff}} = \{x_t - x_{t-1}\}_{t=1}^T, \quad x\in\mathbb{R}^{V\times C}
\end{equation}
for \(t=1,...,T\), where \(T\) is the total number of frames, \(V\) is the number of vertices (joints), and \(C\) represents the coordinates (x and y).
We then compute the Euclidean distance on the difference \(X_{\text{diff}}\) as:
\begin{equation}
    \left \|X_\text{diff}(t,v)\right \|=\sqrt{\sum_{c=1}^{C}(X_{\text{diff}}(t,v,c))^2},
\end{equation}
where \(X_{\text{diff}}(t,v,c)\) denotes the difference at time \(t\) for vertex \(v\) and coordinate \(c\). 
Next, we calculate the mean Euclidean distance across all joints between consecutive frames:
\begin{equation}
    \overline{D}(t)=\frac{1}{V}\sum_{v=1}^{V} \left \|X_\text{diff}(t,v)\right \|.
\end{equation}

Frames where \(\overline{D}(t)\) exceeds a predefined threshold, empirically set to 400, are considered noisy and are subsequently removed.

\begin{figure}[t]
    \centering
    \includegraphics[width=0.8\linewidth]{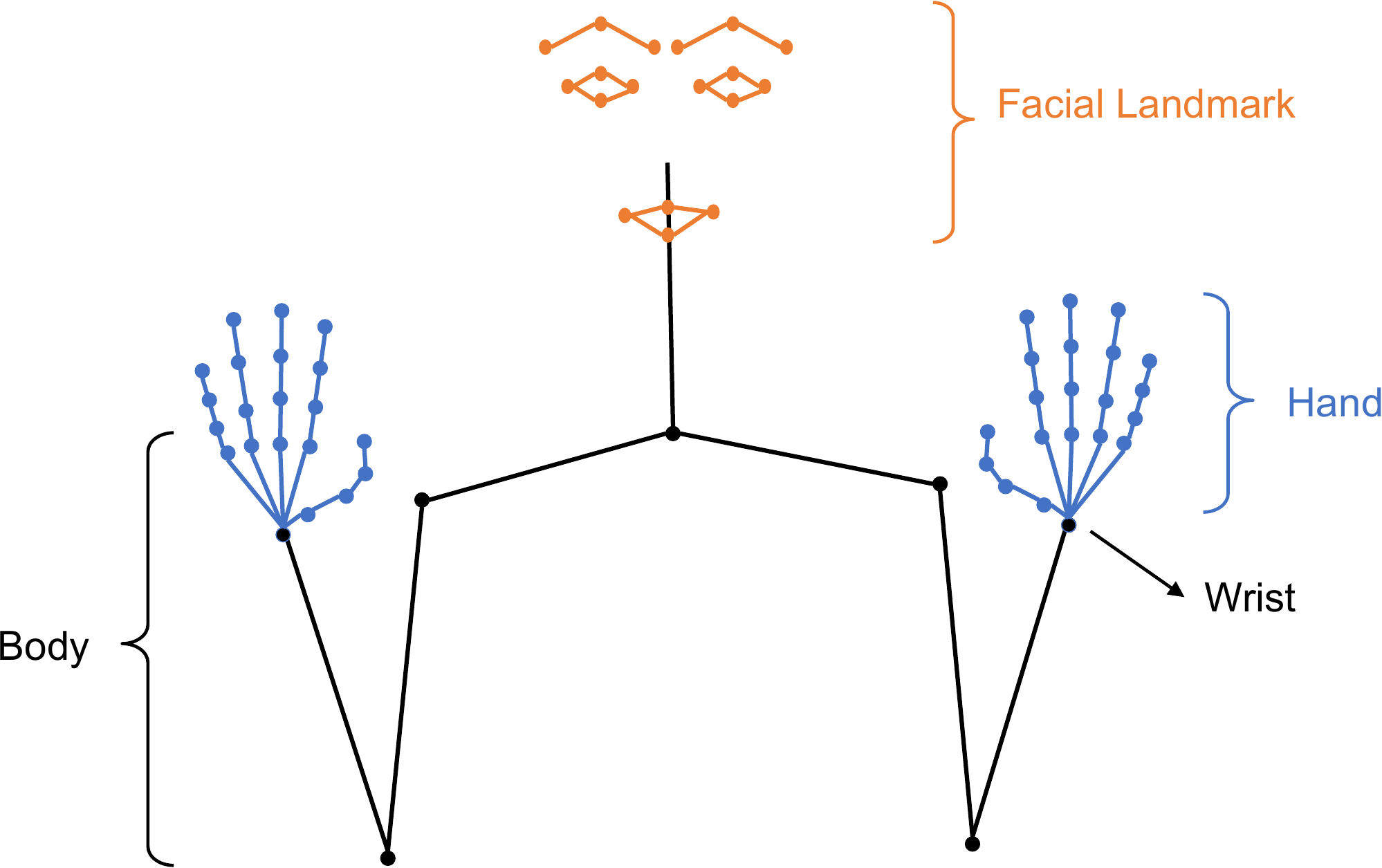}
    \caption{A visual abstract of the keypoints used: 73 keypoints in total, including 19 for the face, 23 for each hand, and 8 for the upper body.}\label{fig:appen_abs_form}
\end{figure}

\begin{figure}[t]
    \centering
    \includegraphics[width=0.6\linewidth,height=0.6\linewidth]{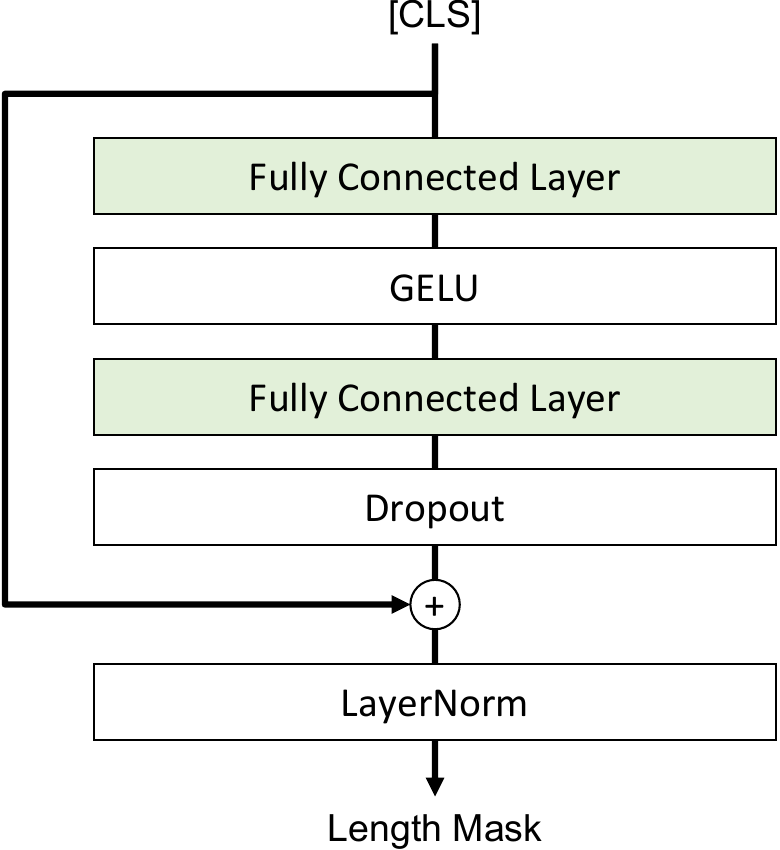}
    \caption{An overview of the length regulator.}
    \label{fig:appen_length_regulator}
    \vspace{-1em}
\end{figure}

\subsection{Network Details}\label{sec:appen_net_details}

We employ a Transformer encoder and decoder architecture~\cite{vaswani2017attention} for our models, largely following the configuration of Joint-SLT~\cite{camgoz2020sign}. Our framework is optimized in two stages: (i) pretraining SignMAE and (ii) task-specific mapping for SLT and SLP. During pretraining, we empirically set a 25\% random mask to extract spatio-temporal features. Typically, MAE mask a large portion of input data (e.g., 75\%) to enhance the robustness of learned representations. However, in our case, where both the encoder and decoder are used for SLT and SLP tasks, the 25\% random mask provides a balanced approach. It offers a sufficient level of masking while retaining enough visible context for effective reconstruction. 

To encode spoken language sentences, we employed DistilBERT\footnote{\scriptsize\url{https://huggingface.co/M-CLIP/M-BERT-Distil-40}}. 
The length regulator module, illustrated in~\cref{fig:appen_length_regulator}, comprises two fully connected linear layers with GELU activation~\cite{hendrycks2016gaussian}, Dropout~\cite{Dropout}, and Layer Normalization~\cite{Ba2016LayerN}. We used a learnable query size of 128 as input to the decoder for generating gloss-level representations. A batch size of 64 was maintained across all tasks, including both pretraining and task-specific mapping. The model was optimized using the AdamW optimizer~\cite{loshchilov2017decoupled} with \(\beta_1=0.9\), \(\beta_2=0.98\), and a weight decay of 1e-3. The learning rate schedule followed a cosine decay, with a peak learning rate of 1e-4 and a linear warmup over 10K steps, decaying to a minimum learning rate of 5e-5.

We trained SignMAE for 50 epochs on a single NVIDIA A100 GPU, completing the process in under 24 hours. 
For sign-to-text mapping, the model was trained for 75 epochs, with early stopping applied when no further improvement in BLEU-4 was observed.
Training took approximately 1 hour for PHOENIX14T and 3 hours for How2Sign. 
We used a beam size of 5.
For text-to-sign mapping, training was completed in 100 epochs, with early stopping based on back-translated BLEU-4. The process was completed in 1 hour for PHOENIX14T and 7 hours for How2Sign.

\section{More Results}\label{sec:appen_more_res}

\subsection{Visualization of the Attention Map}
\begin{figure}[t]
  \centering
  \begin{subfigure}{0.31\linewidth}
    \includegraphics[trim=4cm 3cm 3cm 3cm,clip=true,width=\textwidth,height=5em]{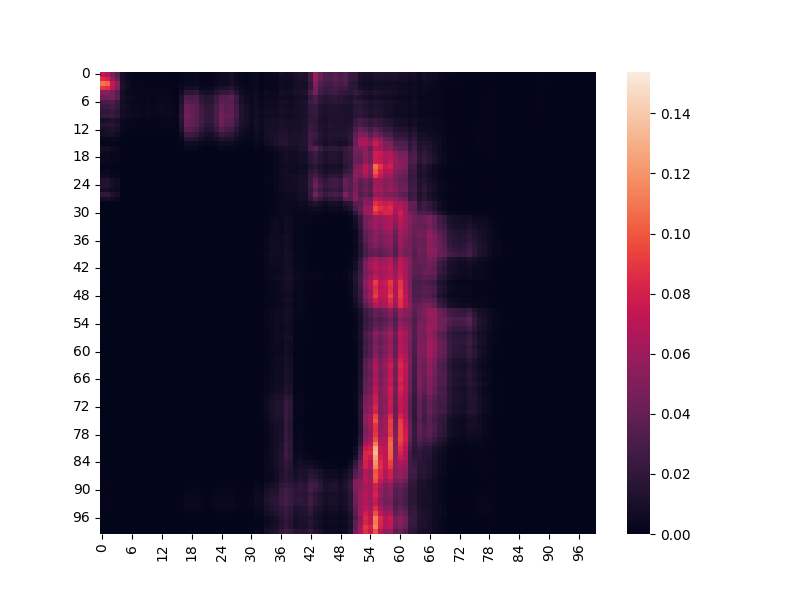}
    \caption{W/ Gloss}\label{fig:abl_attn_with_gloss}
  \end{subfigure}
  \hfill
  \begin{subfigure}{0.31\linewidth}
    \includegraphics[trim=4cm 3cm 3cm 3cm,clip=true,width=\textwidth,height=5em]{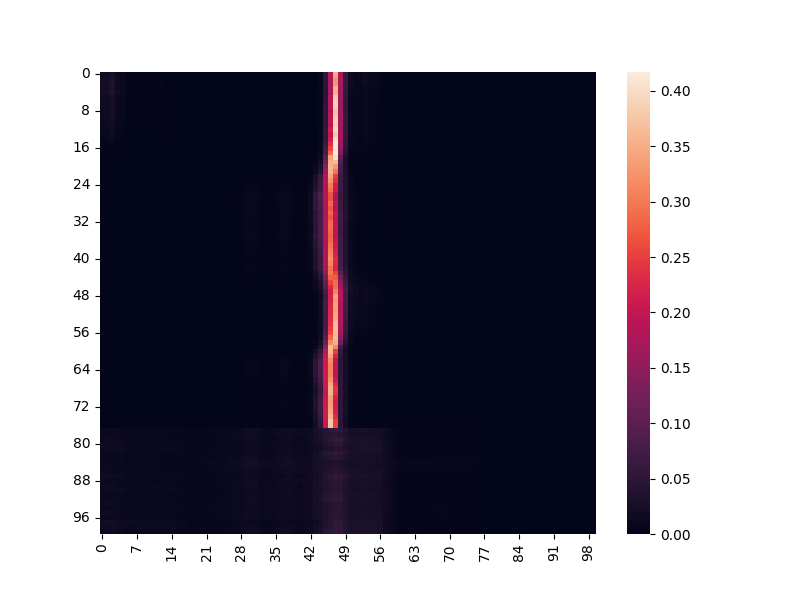}
    \caption{W/o Gloss}\label{fig:abl_attn_without_gloss}
  \end{subfigure}
  \hfill
  \begin{subfigure}{0.31\linewidth}
    \includegraphics[trim=4cm 3cm 3cm 3cm,clip=true,width=\textwidth,height=5em]{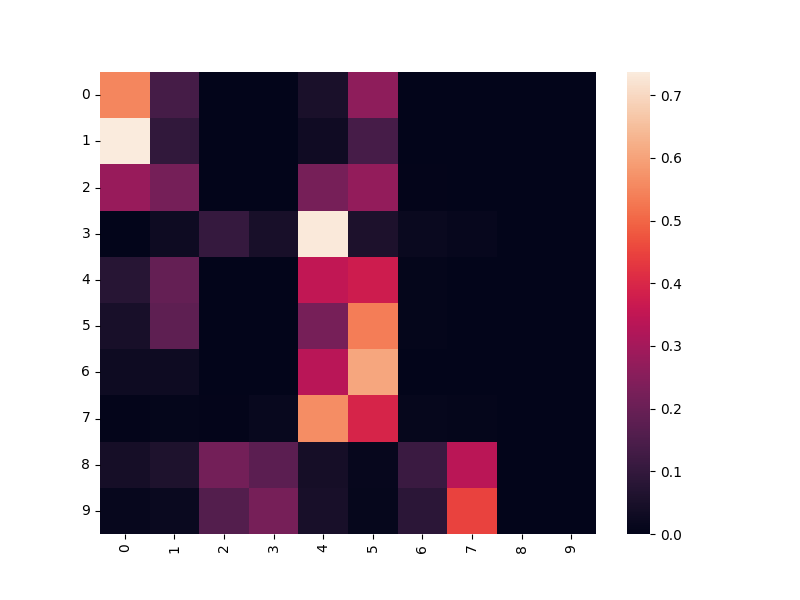}
    \caption{Ours}\label{fig:abl_attn_ours}
  \end{subfigure}
  \caption{Visualization of attention maps in the shallow encoder layer of Joint-SLT with and without glosses, alongside our method. The comparison highlights how each model focuses on different aspects of the input data.}\label{fig:abl_attn}
  \vspace{-1em}
\end{figure}

Building on the findings of~\cite{yin2023gloss}, which demonstrated that glosses provide alignment information, we present visualizations of the attention maps from three different SLT methods: Joint-SLT with and without the glosses, and our method. 
As shown in~\cref{fig:abl_attn_with_gloss}, the use of the glosses clearly helps the model focus on more important local areas by providing essential alignment information. By contrast, as shown in~\cref{fig:abl_attn_without_gloss}, removing the gloss supervision signal causes the model to struggle in identifying the correct regions. However, as shown in~\cref{fig:abl_attn_ours}, our method maintains attention on key regions effectively, performing similarly to the model using the glosses, thereby showing its potential as a gloss replacement.

\subsection{Translation Results}\label{sec:appen_more_trans_res}
We provide additional translation examples in~\cref{tab:more_slt_res}. The results demonstrate that our method consistently delivers accurate and semantically correct translations, while other baselines struggle to capture the correct semantic meaning. 

\subsection{Production Results}\label{sec:appen_more_prod_res}
\begin{figure*}[p]
    \centering
    \begin{subfigure}{\linewidth}
        \includegraphics[width=\linewidth]{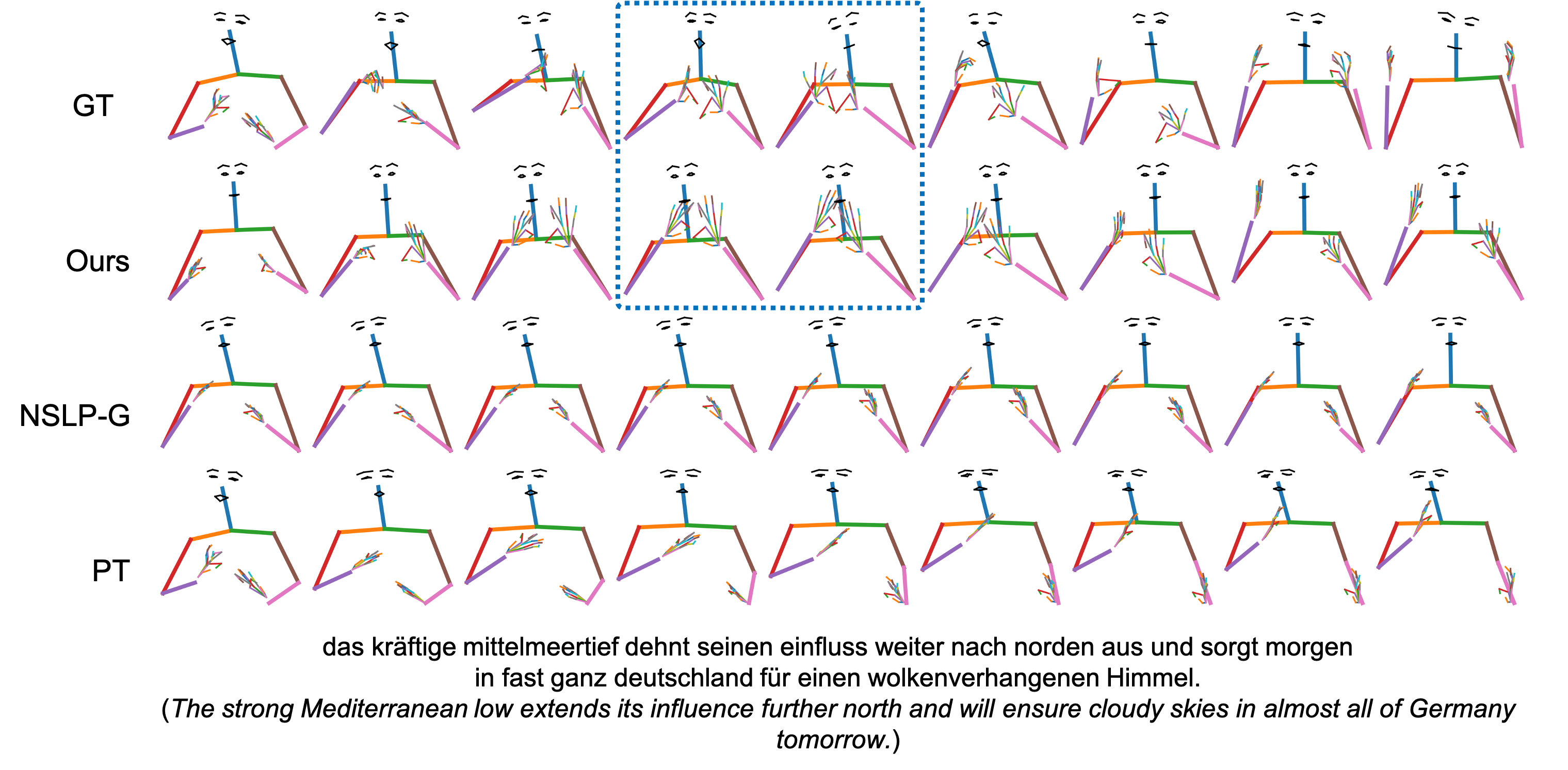}    
    \end{subfigure}
    \vfill\vspace{1em}
    \begin{subfigure}{\linewidth}
        \includegraphics[width=\linewidth]{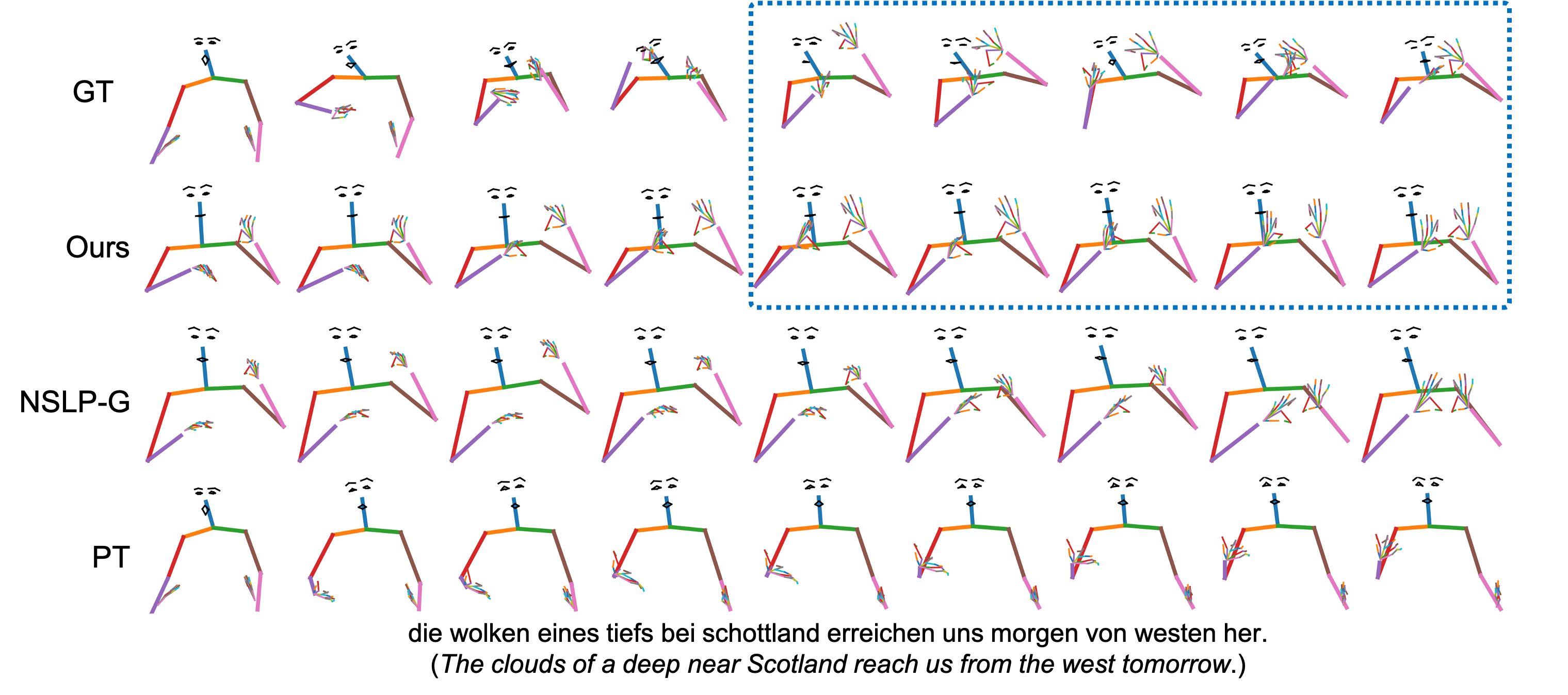} 
    \end{subfigure}
    \caption{Additional visualization examples generated by our method, NSLP-G, and PT. Frames were uniformly selected, with dashed boxes highlighting areas where our method produced more accurate signs.}
    \label{fig:appen_qual_res_p14t_1}
\end{figure*}

\begin{figure*}[!]
    \centering
    \begin{subfigure}{\linewidth}
        \includegraphics[width=\linewidth]{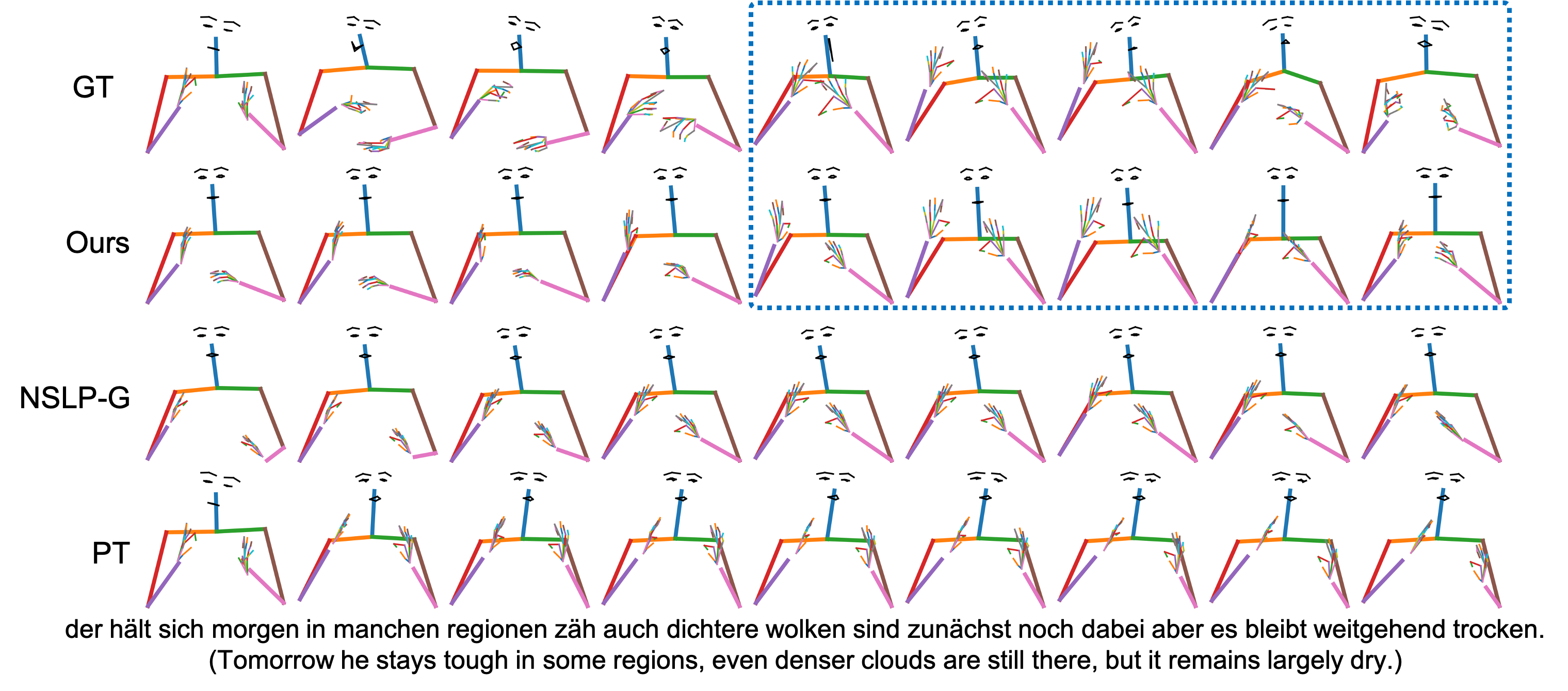}    
    \end{subfigure}
    \vfill\vspace{1em}
    \begin{subfigure}{\linewidth}
        \includegraphics[width=\linewidth]{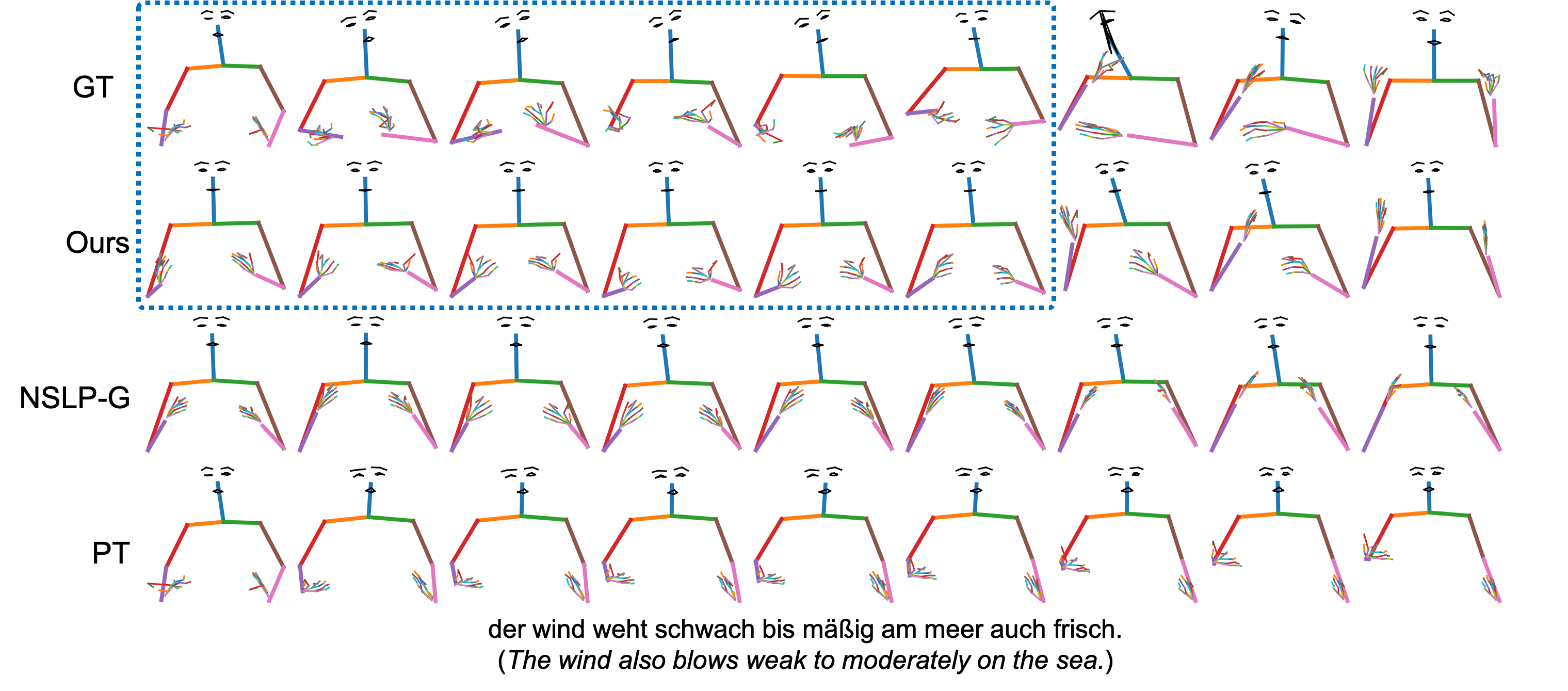} 
    \end{subfigure}
    \caption{Additional visualization examples.}
    \label{fig:appen_qual_res_p14t_2}
\end{figure*}

We provide additional production examples in Figs.~\ref{fig:appen_qual_res_p14t_1} and~\ref{fig:appen_qual_res_p14t_2}. The results demonstrate that our method consistently delivers accurate and semantically correct signs, while other baselines struggle to produce correct signs. 

\section{Potential Negative Societal Impact}\label{sec:social_impact}

On the positive side, the development and refinement of frameworks such as UniGloR, which aim to replace traditional glosses, offer transformative potential, particularly for advancing machine learning applications for Deaf and Hard-of-Hearing communities. By applying this framework to bidirectional sign language translation systems, we can bridge the communication gap between the hearing and Deaf communities. 
This breakthrough holds the promise of creating more inclusive educational, professional, and social environments, reducing the marginalization of Deaf and Hard-of-Hearing individuals.
However, a potential negative impact arises from the datasets we use. Publicly available datasets, such as PHOENIX14T~\cite{camgoz2018neural} and How2Sign~\cite{duarte2021how2sign}, may contain identifiable information, raising concerns about personal privacy. To mitigate these concerns, we work with extracted keypoints from sign videos, ensuring that no personally identifiable information is included in our training process.

\begin{table*}[p]
\scriptsize
\centering
\resizebox{0.8\linewidth}{!}{%
\begin{tabular}{rl} \toprule
    \multicolumn{2}{c}{PHOENIX14T}\\ \midrule 
    Ground Truth:&  und nun die wettervorhersage für morgen montag den achtundzwanzigsten november.\\
    &(\textit{And now the weather forecast for tomorrow monday the twenty-eighth of november.})\\
    Joint-SLT~\cite{camgoz2020sign}&  \textcolor{blue}{und nun die wettervorhersage für morgen} donnerstag \textcolor{blue}{den} sechsundzwanzigsten juli.\\
    &(\textit{And now the weather forecast for tomorrow thursday twenty-sixth of July.})\\  
    ConSLT~\cite{fu2023token}&  \textcolor{blue}{und nun die wettervorhersage für morgen} samstag \textcolor{blue}{den} sechsundzwanzigsten juli.\\ 
    &(\textit{And now the weather forecast for tomorrow saturday the twenty-sixth of July.})\\  
    Ours:&  \textcolor{blue}{und nun die wettervorhersage für morgen} donnerstag \textcolor{blue}{den} siebenundzwanzigsten \textcolor{blue}{november.}\\
    &(\textit{And now the weather forecast for tomorrow thursday the twenty-seventh of November.}) \\ \midrule
    
    Ground Truth:& am samstag ist es in der südhälfte freundlich in der nordhälfte einzelne schauer.\\
    &(\textit{On Saturday, it will be friendly in the southern half and a few showers in the northern half.}) \\
    Joint-SLT~\cite{camgoz2020sign}&\textcolor{blue}{am samstag ist es} im nordwesten \textcolor{blue}{freundlich}\\
    &(\textit{On Saturday it is friendly in the northwest.}) \\
    ConSLT~\cite{fu2023token}&auch \textcolor{blue}{am samstag ist es} im südosten \textcolor{blue}{freundlich.}\\
    &(\textit{Even on Saturday it is friendly in the southeast.}) \\
    Ours:& \textcolor{blue}{am samstag} \textcolor{teal}{scheint im süden oft die sonne und wolken sonst einzelne schauer.}\\
    &(\textit{On Saturday, the sun often shines in the south and clouds otherwise a few showers.}) \\ \midrule
    
    Ground Truth:& am sonntag ziehen von nordwesten wieder schauer und gewitter heran.\\
    &(\textit{On Sunday, showers and thunderstorms will move in again from the northwest.})\\
    Joint-SLT~\cite{camgoz2020sign}& \textcolor{blue}{am} samstag neun grad an der küste.\\
    &(\textit{On Saturday, nine degrees on the coast.})\\
    ConSLT~\cite{fu2023token}& \textcolor{blue}{am sonntag ziehen} dann \textcolor{blue}{von} westen \textcolor{blue}{wieder} \textcolor{teal}{neue niederschläge} \textcolor{blue}{heran.}\\
    &(\textit{On Sunday, new precipitation will move in from the west.}) \\
    Ours:& \textcolor{blue}{am sonntag} \textcolor{teal}{zieht} \textcolor{blue}{von nordwesten wieder} \textcolor{teal}{etwas regen} \textcolor{blue}{heran.} \\
    &(\textit{On Sunday, some rain will move in again from the northwest.}) \\ \midrule 

    \multicolumn{2}{c}{How2Sign}\\ \midrule
    Ground Truth:& Today, I'm going to show you how to play three card poker.\\
    Joint-SLT~\cite{camgoz2020sign}& So what we're \textcolor{blue}{going to} do is we're going to take our sponge.\\
    ConSLT~\cite{fu2023token}&You want to make sure that you have your seatbelt buckled.\\
    Ours:&\textcolor{blue}{Today I'm going to show you how to} do this.\\ \midrule
    Ground Truth:&Now we're going to talk specifically about the medication that is going to be used with this machine.\\
    Joint-SLT~\cite{camgoz2020sign}& So what \textcolor{blue}{we're going to} do is we're going to start on the bottom here and we're going to do it in slow-motion.\\
    ConSLT~\cite{fu2023token}& So what \textcolor{blue}{we're going to} do is we're going to start by pruning on the bottom and we're going to go over the top.\\
    Ours:& In this clip, \textcolor{blue}{we're going to talk} about the proper way to get rid of the skin.\\ \bottomrule
    \end{tabular}
}
\caption{Comparison of the translation results compared to baselines. Correctly translated 1-grams are highlighted in \textcolor{blue}{blue}, and semantically correct translations are highlighted in \textcolor{teal}{green}.}\label{tab:more_slt_res}
\end{table*}

\end{document}